\pdfoutput=1
\documentclass[11pt]{article}

\usepackage[final]{acl}

\usepackage{times}
\usepackage{latexsym}
\usepackage[T1]{fontenc}
\usepackage[utf8]{inputenc}
\usepackage{microtype}
\usepackage{inconsolata}

\usepackage{graphicx}
\graphicspath{{./figures/}}
\usepackage{booktabs}
\usepackage{array}

\title{Measuring Concept Content in Text from LLM Activations:\\ ESG Evidence from Concept Vectors and Linear Probes}

\author{Luc Hazenoot \quad Zhaochun Ren \quad Amirhossein Zohrehvand \\
  Leiden Institute of Advanced Computer Science, Leiden University}

\begin{document}
\maketitle

\begin{abstract}
Recent work shows that Large Language Models (LLMs) encode more information in their activations than they show in their responses. We ask whether the information contained in activations could be used as a source of measurement for the presence of concepts. We compare two algorithms for extraction of concepts, i.e., Recursive Feature Machine (RFM) algorithm and linear probing against an embedding baseline, surface baselines, and the model's own response using a human-annotated Environmental, Social and Governance (ESG) dataset of financial texts.
The best linear probe ranges from 0.6 to 3.3 percentage points below fine-tuned domain classifiers. The probes also outscore the model's own response in all comparisons.
Furthermore, the probes outperform the RFM concept vectors in all accuracy comparisons. In a synthetic mixture test, their mean-pooled score tracks the number of concept sentences in a paragraph better than the concept-vector score. Put together, a linear probe on a frozen LLM is a cost-effective and generalizable alternative to fine-tuning a domain classifier.
\end{abstract}

\section{Introduction}
Recent work shows that Large Language Models (LLMs) encode more information in their activations than they show in their responses \citep{doi:10.1126/science.aea6792,burns2023discovering,azaria-mitchell-2023-internal,li2023inference,marks2024geometry,gurnee2026workspace}.
The linear representation hypothesis proposes that concepts are linearly accessible as directions in activation space \citep{pmlr-v80-kim18d,pmlr-v235-park24c,marks2024geometry}, and therefore can be extracted and used for downstream tasks.
Building on this theory, \citet{doi:10.1126/science.aea6792} extract such directions as concept vectors with the Recursive Feature Machine (RFM) algorithm \citep{radhakrishnan2024mechanism} and use them to steer and monitor models across hundreds of concepts.
\citet{yang2024llmmeasure} project the activations of a frozen LLM onto a learned concept direction to obtain text-based measures for social science research, validated by their correlation with downstream outcomes.

Most text-based measures, however, use the surface of a text or model responses, and not the activations.
Dictionaries count concept words, topic models allocate words to topics, and embedding methods compare texts to a concept direction in embedding space \citetext{\citealp{loughran2011liability,blei2003latent,gentzkow2019text,kozlowski2019geometry,li2021measuring}; for a recent review of NLP work in the field of finance see \citealp{DU2025102755}}.
The latest practices build on advances in LLMs, using the model's own response to a prompt to measure concepts out of text \citep{gilardi2023chatgpt,rathje2024gpt,ziems2024can}.
What is less understood is whether the information contained in the activations helps with improving the measurement of concepts compared to relying on the model's own response.

The purpose of this paper is to test whether the information contained in the LLM activations, in the form of internal concept vectors, can be used to measure concepts from texts, and compare it to the model's own response.
We compare two algorithms for extraction of concepts (extractors), the RFM algorithm and linear probing, on four frozen LLMs against an embedding baseline, surface baselines, the model's own response, and the fine-tuned domain classifiers of a human-annotated Environmental, Social and Governance (ESG) dataset of financial texts \citep{schimanski2024bridging}.
We find that the linear probe outscores the model's own response in all twelve comparisons, for each of the four models, indicating that the activations contain information that the model does not show in its response.
Overall, the probe reaches the accuracy of the model's own response in eleven of the twelve experimental cells.
Without any task-specific fine-tuning, the best linear probe is 0.6, 1.2 and 3.3 percentage points below the fine-tuned domain classifiers on the Environmental, Social and Governance pillars, respectively.
The simpler linear probe was the best activation-based method for this particular dataset.
In a synthetic mixture test, the probe's mean-pooled score tracks the number of concept sentences in a paragraph better than the concept-vector score. Current ESG datasets carry only binary labels, so validation against human-graded labels to measure concept degree is left to future work.
So far, the gap between activations and responses has been shown for truthfulness and hallucination \citep{liu-etal-2023-cognitive,orgad2025llms}, while ESG and climate text is still measured with fine-tuned domain classifiers \citep{webersinke2021climatebert,bingler2022cheap,stammbach-etal-2023-environmental,schimanski-etal-2023-climatebert}.
Our contribution is twofold. First, we show that the information contained in the activations outscores the model's own response as a source of measurement for the presence of concepts. Second, we demonstrate that leveraging the internal representation of an LLM can be a cost-effective and generalizable alternative to fine-tuning a domain classifier.

\section{Background and Related Work}
Measuring a concept from a frozen model rests on a premise and two design choices.
The premise is that the concept is present in the model's activations as a direction that can be read off them.
The choices are how to turn labelled activations into that direction, and how to reduce the many activation vectors a sentence produces into the single vector both extractors require. The experiments vary both.
Appendix \ref{app:background} gives background on activations, concept vectors, linear probing, RFM and token pooling.

Linear probing uses lightweight linear classifiers to interpret the activation space of a frozen LLM for classification tasks \citep{DBLP:conf/iclr/AlainB17,10.1162/coli_a_00422}.
The classifier is trained on labelled activations to find the direction that best separates the desired concept from its absence.
Its prediction is then interpreted as how strongly the concept is represented in that layer \citep{10.1162/coli_a_00422,hewitt-liang-2019-designing}.
Common models chosen include linear ridge regression \citep{Hoerl01021970} and logistic regression.

\section{Methods}

\subsection{Baseline and Models}
As a baseline, we use an embedding model.
Embedding models are a standard way to turn text into dense feature vectors that can be used for classification \citep{reimers-gurevych-2019-sentence}.

For the embedding baseline, we use the Qwen-3-Embedding-8B model.
We choose this model because it is a recent embedding model that is based on the Qwen-3 family that we also use for our activation-based experiments.
We follow the model's recommended input format, prefixing every sentence with a task instruction that describes the pillar being classified, so that the baseline receives the same kind of task framing that the wrapper provides to the activation-based methods.
We take the resulting embeddings and use them as input features in a logistic regression model to predict the binary labels.
We do the prediction by taking the mean of the metrics over a 5-fold stratified cross-validation (seed 23).
We standardise the embeddings, use balanced class weights, and select the regularisation strength $C$ from six values by mean AUC over the same folds, which makes this baseline slightly optimistic.

For our experiments, we used a range of decoder-only models to test how different families and sizes perform: Llama-3.1-8B, Qwen-3-8B, Qwen-3-14B and Gemma-4-31B-it (Appendix \ref{app:impl}), next to the Qwen-3-Embedding-8B baseline.
The own-answer baseline asks each model a yes-or-no question about the pillar (prompts in Table \ref{tab:wrappers_used}, details in Appendix \ref{app:impl}).
We sum the probabilities of the single-token variants of ``yes'' (yes, Yes and YES, with and without a leading space), do the same for ``no'', and score each sentence by $P(\mathrm{Yes}) / (P(\mathrm{Yes}) + P(\mathrm{No}))$.
AUC uses this score without a threshold, and accuracy predicts the concept when $P(\mathrm{Yes}) > P(\mathrm{No})$.
The surface baselines fit a logistic regression, under the same 5-fold stratified cross-validation, to the share of a sentence's tokens that match a 16-term concept list per pillar (dictionary), to 25 LDA topic proportions, or to TF-IDF word 1- and 2-grams.

\subsection{Dataset}
The dataset we use is a human-annotated ESG dataset by \citet{schimanski2024bridging}, containing 2000 sentences per ESG pillar: Environmental, Social, Governance.
The source is a corpus of 13.8 million English sentences extracted from: annual reports, responsibility reports, sustainability reports and news articles.
Each dataset is built out of the following: 1000 shared sentences, 750 domain-specific sentences and 250 general language sentences.
Each sentence contains a binary label: 0 if the domain is not present and 1 if it is.

The 1000 sentences are shared between the three datasets, and each sentence is annotated three times for its respective pillar.
A shared sentence, therefore, might contain multiple positive labels.

The 750 domain-specific sentences are subdivided into two categories: 500 sentences filtered by the pillar keywords and 250 with no keyword filter.
Sentences filtered by the pillar keywords are not necessarily true positives.
A sentence containing the keyword ``green'', for example, could be about a traffic light rather than the environment.

The last 250 general sentences are unfiltered sentences from the corpus, mostly containing negative sentences that are unrelated to the specific pillar.
Three expert annotators then labelled each sentence following a shared rubric.
Table \ref{tab:iaa} (Appendix \ref{app:impl}) shows their inter-annotator agreement.

Table \ref{tab:label_dist} (Appendix \ref{app:impl}) shows the resulting label distribution per dataset.

\subsection{Experiments}
\subsubsection{RFM}
Our primary experiments use the RFM algorithm introduced by \citet{radhakrishnan2024mechanism}, in the xRFM implementation of \citet{beaglehole2026xrfm}.
We first do a random shuffle on the dataset, and then divide it into three splits: a training split of 1000 sentences, a validation split of 500 sentences, and a test split of 500 sentences.
Every sentence is wrapped in a fixed statement whose purpose is to guide the model towards the concept we want to extract. Finally, the sentences are provided with a chat template as the user's turn.

Using the training split, we extract the last token activations of each layer of the model and use them as input to the RFM algorithm.
From the resulting AGOP matrix, we take the top $N = 5$ eigenvectors by eigendecomposition and use them as our concept vectors.
Because the sign of an eigenvector is arbitrary, each vector is oriented before use: within every layer we project the training activations onto the vector and flip its sign whenever that projection correlates negatively with the labels.
Without this step, the per-vector scores would not be commensurable and could not be averaged.

To score a new sentence, we compute the cosine similarity between its activations and the concept vectors.
For each concept vector $n$, we first compute the per-layer cosine similarity between the activation and the concept vector, and then average these per-layer values into a single score for that concept vector.
This is repeated for all $N$ concept vectors, and the $N$ resulting scores are averaged into one final score per sentence.
We compute this final score for every sentence in the split. For display only (Appendices B and I), the scores are min-max scaled within each pillar and split to fit between $-1$ and $1$.
An optimal threshold is then found on the raw, unscaled scores in order to predict the label.
This threshold is fitted on the validation split and applied unchanged to the test split, which is also the split the paired statistics use (Section \ref{sec:stats}).
Sweeping it on the test split instead raises accuracy by 0.013 on average across the twenty-four cells, and F1 by 0.021, with worst cases of 0.038 and 0.252 (Table \ref{tab:rfm_swept}, Appendix \ref{app:threshold}).
Accuracy is affected little. F1 is not, because the sweep maximises accuracy and the F1 at that operating point is incidental to it.
From this we calculate the following metrics: Area Under the Curve (AUC), accuracy, F1, precision and recall.

Because concept vectors differ based on what wrapper was used, we evaluate the wrappers against each other based on accuracy.
This is because \citet{schimanski2024bridging} evaluated their baseline models on this metric, which gives us a direct comparison against them.
Wrappers were chosen by hand during development from two to nine candidates per model and pillar. For RFM, we selected the wrapper by the validation accuracy of the mean of the top five concept vectors, and used the same run for the first concept vector and the model's own answer. During development, probe wrappers were compared on 5-fold cross-validation accuracy over all 2,000 sentences (Appendix \ref{app:impl}). Every reported number comes from a rerun with the chosen wrapper per method (Table \ref{tab:wrappers_used}), so the probe and RFM use different wrappers in ten of the twelve model-pillar cells.

The concept vectors are the top eigenvectors of the AGOP matrix, ordered by eigenvalue, so the first concept vector is the one that separates the classes the strongest.
We report the metrics of the first concept vector on its own, alongside the mean of the top 5, to see how much of the performance comes from this single strongest direction compared to using multiple concept vectors together.

We also show the continuous scaled scores and their distribution.
These scores reflect the similarity between the concept vectors and new sentences, where a higher score corresponds to a stronger alignment with the concept.
This score is intended to reflect how much the concept is present in a new sentence, rather than only whether it is present.

\subsubsection{Linear Probing}
For the linear probing approach, we use a stacked generalisation architecture.
In the first stage, a separate RidgeClassifier with strong L2 regularisation is trained on each layer's activations.
We probe every hidden layer except the earliest one, giving 31 layers for Llama-3.1-8B, 35 for Qwen-3-8B, 39 for Qwen-3-14B and 59 for Gemma-4-31B-it.
This produces a per-layer decision score.
In the second stage, a logistic regression meta-learner is trained on the aggregated per-layer scores to produce a final prediction.
To avoid any data leakage when training the logistic model, we use a nested cross-validation procedure.
The outer loop is a 5-fold stratified cross-validation used to compute the final score.
Within each outer training fold, an inner 5-fold stratified cross-validation generates out-of-fold predictions.
This means that each training example receives a decision score from a probe that was not exposed during fitting.
These out-of-fold scores serve as the logistic regression model's training features.
Inner fold probes are then refit on the full outer training fold to produce decision scores on the outer test set.
Standardisation is also performed inside each respective inner and outer fold so that scaling statistics are computed only on the relevant training slice in both the inner and outer loops to avoid minor data leakage.
The hyperparameters remain fixed: ridge $\alpha = 200$, meta-learner $C = 0.1$, decision threshold 0.5, and inner and outer 5-fold stratified cross-validation with seed 992.

Since each layer of an LLM produces one activation vector per input token, a pooling strategy is required to obtain a single activation per layer that best represents the information learned in that respective layer.
We compare three strategies: last-token, mean pooling, and max pooling.

\subsection{Comparison of methods}
Both linear probing and RFM are supervised methods that aim to find a direction in the activation space that best separates prompts where the target concept is present from those where it is not.
Linear probing trains a linear classifier on the activations. The direction normal to its decision boundary is the direction that best separates the two classes \citep{10.1162/coli_a_00422}.
RFM instead trains a non-linear predictor on the activations and extracts the top eigenvectors of the AGOP matrix, which are taken as the concept vectors.
To detect the concept in a new prompt, these per-layer concept vectors are projected onto the prompt's activations at each layer, and the resulting similarities are aggregated into a single score that is thresholded to make the prediction.

\subsection{Statistics and additional tests}
\label{sec:stats}
The activation-based and own-answer cells carry 95\% percentile bootstrap confidence intervals, computed from 10,000 resamples of the sentences with seed 0.
Two probe protocols appear: the CV probe, scored by 5-fold cross-validation over all 2,000 sentences (accuracy and AUC in Table \ref{tab:main}), and the split probe, trained on the 1,000 training sentences and tested on the 500 test sentences (paired tests, label efficiency and mixture test). Both use the primary pooling, the pooling with the highest validation accuracy for each model and pillar, with ties broken in the fixed order last token, max, mean, unless another pooling is named, and the primary probe of the mixture test is the split probe at that pooling.
Paired bootstrap and McNemar tests, Holm-corrected over the twelve model-pillar cells, compare the split probe with the own answer and with RFM on the 500 test sentences.
Label-efficiency curves refit the split probe on subsets of the training sentences, and the mixture test (Figure \ref{fig:mixture}) scores synthetic paragraphs with $k$ concept sentences out of $n$ and reports Spearman's $\rho$ with $k$.
Appendix \ref{app:stats} gives the details.

\section{Results}
Table \ref{tab:main} reports accuracy and AUC for every method on the Environmental, Social and Governance pillars, and Tables \ref{tab:full_environmental}, \ref{tab:full_social} and \ref{tab:full_governance} in Appendix \ref{app:full} report every metric.

\begin{table*}[t]
  \centering
  \small
  \begin{tabular}{@{}l@{\hspace{8pt}}l@{\hspace{10pt}}c@{\hspace{8pt}}c@{\hspace{12pt}}c@{\hspace{8pt}}c@{\hspace{12pt}}c@{\hspace{8pt}}c@{}}
\toprule
 & & \multicolumn{2}{c}{Environmental} & \multicolumn{2}{c}{Social} & \multicolumn{2}{c}{Governance} \\
\cmidrule(lr){3-4} \cmidrule(lr){5-6} \cmidrule(lr){7-8}
Method & Model & Acc & AUC & Acc & AUC & Acc & AUC \\
\midrule
Own answer & Gemma-4-31B-it & 0.930{\scriptsize $\pm$0.023} & 0.973{\scriptsize $\pm$0.013} & 0.804{\scriptsize $\pm$0.035} & 0.860{\scriptsize $\pm$0.037} & 0.756{\scriptsize $\pm$0.037} & 0.834{\scriptsize $\pm$0.037} \\
 & Llama-3.1-8B & 0.864{\scriptsize $\pm$0.029} & 0.969{\scriptsize $\pm$0.013} & 0.822{\scriptsize $\pm$0.033} & 0.892{\scriptsize $\pm$0.028} & 0.820{\scriptsize $\pm$0.033} & 0.842{\scriptsize $\pm$0.040} \\
 & Qwen-3-8B & 0.920{\scriptsize $\pm$0.023} & 0.980{\scriptsize $\pm$0.010} & 0.836{\scriptsize $\pm$0.032} & 0.937{\scriptsize $\pm$0.019} & 0.842{\scriptsize $\pm$0.032} & 0.830{\scriptsize $\pm$0.047} \\
 & Qwen-3-14B & 0.916{\scriptsize $\pm$0.025} & 0.978{\scriptsize $\pm$0.010} & 0.810{\scriptsize $\pm$0.035} & 0.893{\scriptsize $\pm$0.028} & 0.800{\scriptsize $\pm$0.035} & 0.861{\scriptsize $\pm$0.038} \\
\midrule
RFM first & Gemma-4-31B-it & 0.936{\scriptsize $\pm$0.021} & 0.975{\scriptsize $\pm$0.013} & 0.868{\scriptsize $\pm$0.029} & 0.943{\scriptsize $\pm$0.019} & 0.770{\scriptsize $\pm$0.038} & 0.811{\scriptsize $\pm$0.044} \\
 & Llama-3.1-8B & 0.894{\scriptsize $\pm$0.027} & 0.962{\scriptsize $\pm$0.015} & 0.852{\scriptsize $\pm$0.031} & 0.922{\scriptsize $\pm$0.023} & 0.842{\scriptsize $\pm$0.031} & 0.877{\scriptsize $\pm$0.036} \\
 & Qwen-3-8B & 0.892{\scriptsize $\pm$0.027} & 0.965{\scriptsize $\pm$0.013} & 0.818{\scriptsize $\pm$0.034} & 0.907{\scriptsize $\pm$0.026} & 0.824{\scriptsize $\pm$0.033} & 0.863{\scriptsize $\pm$0.038} \\
 & Qwen-3-14B & 0.936{\scriptsize $\pm$0.021} & 0.984{\scriptsize $\pm$0.008} & 0.854{\scriptsize $\pm$0.031} & 0.935{\scriptsize $\pm$0.020} & 0.846{\scriptsize $\pm$0.031} & 0.872{\scriptsize $\pm$0.039} \\
\midrule
RFM mean of 5 & Gemma-4-31B-it & 0.940{\scriptsize $\pm$0.021} & 0.983{\scriptsize $\pm$0.009} & 0.834{\scriptsize $\pm$0.032} & 0.922{\scriptsize $\pm$0.022} & 0.810{\scriptsize $\pm$0.035} & 0.855{\scriptsize $\pm$0.039} \\
 & Llama-3.1-8B & 0.872{\scriptsize $\pm$0.029} & 0.943{\scriptsize $\pm$0.020} & 0.824{\scriptsize $\pm$0.034} & 0.902{\scriptsize $\pm$0.027} & 0.840{\scriptsize $\pm$0.032} & 0.881{\scriptsize $\pm$0.033} \\
 & Qwen-3-8B & 0.844{\scriptsize $\pm$0.033} & 0.912{\scriptsize $\pm$0.024} & 0.768{\scriptsize $\pm$0.037} & 0.841{\scriptsize $\pm$0.034} & 0.802{\scriptsize $\pm$0.035} & 0.847{\scriptsize $\pm$0.039} \\
 & Qwen-3-14B & 0.938{\scriptsize $\pm$0.021} & 0.980{\scriptsize $\pm$0.009} & 0.802{\scriptsize $\pm$0.035} & 0.910{\scriptsize $\pm$0.025} & 0.810{\scriptsize $\pm$0.034} & 0.871{\scriptsize $\pm$0.034} \\
\midrule
Linear probe & Gemma-4-31B-it & 0.949{\scriptsize $\pm$0.010} & 0.984{\scriptsize $\pm$0.005} & 0.915{\scriptsize $\pm$0.012} & 0.973{\scriptsize $\pm$0.006} & 0.861{\scriptsize $\pm$0.015} & 0.911{\scriptsize $\pm$0.014} \\
 & Llama-3.1-8B & 0.945{\scriptsize $\pm$0.010} & \textbf{0.985}{\scriptsize $\pm$0.005} & 0.921{\scriptsize $\pm$0.012} & 0.972{\scriptsize $\pm$0.007} & 0.860{\scriptsize $\pm$0.016} & 0.913{\scriptsize $\pm$0.014} \\
 & Qwen-3-8B & \textbf{0.951}{\scriptsize $\pm$0.010} & 0.983{\scriptsize $\pm$0.006} & 0.922{\scriptsize $\pm$0.012} & 0.976{\scriptsize $\pm$0.006} & 0.859{\scriptsize $\pm$0.015} & 0.909{\scriptsize $\pm$0.015} \\
 & Qwen-3-14B & 0.943{\scriptsize $\pm$0.010} & 0.982{\scriptsize $\pm$0.006} & 0.920{\scriptsize $\pm$0.012} & 0.976{\scriptsize $\pm$0.006} & \textbf{0.863}{\scriptsize $\pm$0.015} & 0.910{\scriptsize $\pm$0.014} \\
\midrule
Embedding + LR & Qwen-3-Emb-8B & 0.943{\scriptsize $\pm$0.005} & 0.981{\scriptsize $\pm$0.007} & \textbf{0.925}{\scriptsize $\pm$0.019} & \textbf{0.977}{\scriptsize $\pm$0.004} & 0.836{\scriptsize $\pm$0.007} & \textbf{0.918}{\scriptsize $\pm$0.012} \\
\midrule
TF-IDF + LR & 1,2-grams & 0.883{\scriptsize $\pm$0.017} & 0.954{\scriptsize $\pm$0.011} & 0.865{\scriptsize $\pm$0.024} & 0.943{\scriptsize $\pm$0.008} & 0.856{\scriptsize $\pm$0.008} & 0.894{\scriptsize $\pm$0.018} \\
\midrule
Fine-tuned & RoBERTa & \textbf{0.957}{\scriptsize $\pm$0.010} &  & \textbf{0.934}{\scriptsize $\pm$0.014} &  & 0.896{\scriptsize $\pm$0.011} &  \\
 & DistilRoBERTa & 0.950{\scriptsize $\pm$0.011} &  & 0.932{\scriptsize $\pm$0.015} &  & \textbf{0.897}{\scriptsize $\pm$0.011} &  \\
\bottomrule
\end{tabular}

  \caption{Accuracy and AUC on the three ESG pillars. Own answer and RFM (first concept vector, and mean of the top 5, threshold fitted on validation) are scored on the 500-sentence test split, the linear probe (primary pooling) and the embedding and TF-IDF baselines by 5-fold cross-validation over all 2,000 sentences. $\pm$: half-width of the 95\% bootstrap confidence interval, or the sd over folds for the embedding, TF-IDF and fine-tuned rows. Fine-tuned: \citet{schimanski2024bridging}, as published, no AUC reported. Bold: best per pillar and metric, with the fine-tuned rows bolded separately.}
  \label{tab:main}
\end{table*}

\subsection{Activation-based comparison}
Across all pillars, linear probing achieves the highest results of the activation-based methods and comes close to the fine-tuned models from \citet{schimanski2024bridging}.

On the Environmental dataset, the best linear probing experiment, which used last token activation on the Qwen-3-8B model, scored 0.951 accuracy, coming within 0.6 percentage points of the fine-tuned EnvRoBERTa, which achieved 0.957 on accuracy.

On the Social dataset, the baseline embedding method scored 0.925 on accuracy.
The best linear probing method, using max pooling and Qwen-3-8B, trails it by 0.3 points at 0.922, a gap far smaller than the cross-validation standard deviation of the embedding method.
The best fine-tuned model, SocRoBERTa, scored 0.934.

On the Governance dataset, the baseline embedding method scored 0.836 on accuracy.
Using linear probing, with max pooling and the Llama-3.1-8B model, we achieved a score of 0.864 (Table \ref{tab:full_governance}).
The best fine-tuned model, GovDistilRoBERTa, scored 0.897.

That comparison holds when both methods are given the same sentences, each method with its own chosen wrapper (Table \ref{tab:wrappers_used}).
With the wrapper held fixed as well, the Governance sweep in Appendix A puts the probe ahead on accuracy of the first concept vector in twenty of the twenty-one model-wrapper cells and of the mean of five in all twenty-one.
Because linear probing is evaluated by cross-validation over all 2,000 sentences while RFM is evaluated once on a held-out split, the two are not trained on the same material.
We therefore repeated the probe on exactly the split RFM uses: the same 1,000 training sentences and the same 500 test sentences.
Linear probing remains ahead of the first concept vector in all twelve model-pillar combinations, by 4.3 accuracy points on average, so the ordering does not follow from the evaluation protocol.
The difference is Holm-significant in five of the twelve combinations on accuracy and in eight on AUC (Appendix \ref{app:paired}).

For linear probing, there is no single best pooling strategy. Last token pooling is best on Environmental, while max token pooling is best on Social and Governance.
The same holds for model selection on linear probing. Qwen-3-8B performs the best on Environmental and Social, while Llama-3.1-8B performs best on Governance.

For RFM, the first concept vector beats the mean of the top five concept vectors on accuracy in nine of the twelve model-pillar combinations.

\subsection{Comparison with the model's own answer}
We prompted each of the four models with the same concept definitions as RFM on the same test sentences, asked for a yes-or-no verdict, and scored the answer by the probability it assigns to ``Yes'' against ``No'', which yields an AUC directly comparable to the activation-based methods.

The split probe beats the model's own answer in all twelve model-pillar combinations on both accuracy and AUC, by 6.1 accuracy points and 0.046 AUC on average.
The difference is Holm-significant in nine of the twelve on accuracy and in eight on AUC, and the cells that are not are mostly on Environmental, where both methods are near ceiling.
Interpreting the internal activations therefore recovers concept content better than reading the generated response, which is the claim the approach rests on.
The first RFM concept vector clears the same bar on AUC in eight of twelve, and on Environmental with the two smaller models, on Social with Qwen-3-8B and on Governance with Gemma-4-31B-it it falls below the model's response.

Dictionary counts and topic proportions, the surface measures this approach is meant to improve on, sit far below both: a concept-word dictionary reaches 0.128 F1 on Governance against 0.733 for the best probe, and a 25-topic model reaches 0.373.
A tuned bag-of-words classifier is a harder baseline than either.
TF-IDF features with logistic regression reach 0.894 AUC on Governance, above every RFM configuration in Table \ref{tab:full_governance}, though even there they trail the best probe by 0.8 accuracy points (0.856 against 0.864), and on Environmental and Social the gap widens to 6.8 and 5.7 accuracy points.

\subsection{Label efficiency}
\label{sec:labeleff}
Figure \ref{fig:label-efficiency} in Appendix \ref{app:labeleff} shows the split probe's test accuracy against the number of training labels.
With 100 labels, the probe reaches the accuracy of the first RFM concept vector trained on all 1,000 labels in seven of the twelve model-pillar cells (three of them by margins within one sd), and that of the model's own answer in eight of twelve.
By 250 labels, it reaches them in nine and eleven of twelve cells.
Table \ref{tab:label_eff} in Appendix \ref{app:labeleff} gives the numbers.

\subsection{Does the score track how much?}
\label{sec:mixture}
\begin{figure*}[t]
  \centering
  \includegraphics[width=\textwidth]{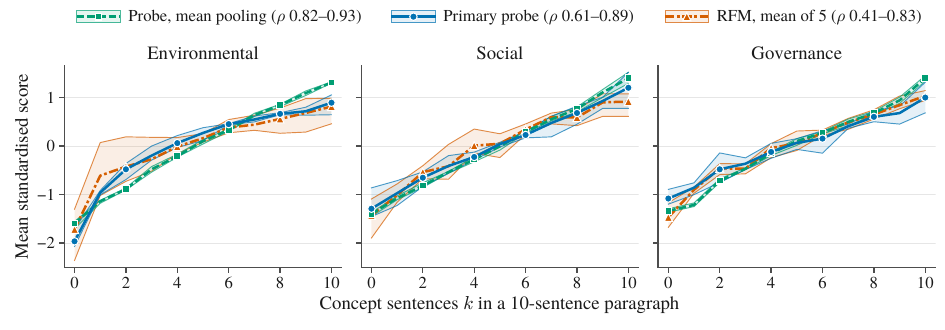}
  \caption{Synthetic mixture test: mean standardised score against the number $k$ of concept sentences in a 10-sentence paragraph (330 paragraphs per pillar, model and method). Each line averages the four models, with a band from the lowest to the highest model. All three rise with $k$ in every pillar.}
  \label{fig:mixture}
\end{figure*}

The synthetic mixture test asks whether a method's continuous score rises with the number $k$ of concept sentences in a paragraph (Figure \ref{fig:mixture}).
At $n = 10$, the Spearman correlation of score with $k$ ranges from 0.82 to 0.93 for the mean-pooled probe, from 0.61 to 0.89 for the primary probe and from 0.41 to 0.83 for the RFM mean of five across the twelve model-pillar cells, and every confidence interval lies above zero.
Cell by cell at $n = 10$, the RFM mean of five falls below the mean-pooled probe in all twelve cells and below the primary probe in seven.
Averaged over the four models, the mean-pooled probe has the highest correlation with $k$ of all methods in every pillar at both paragraph lengths ($n = 5$ and 10).
Table \ref{tab:mixture_rho} in Appendix \ref{app:mixture} gives every cell at $n = 10$.

\section{Discussion}
\label{sec:discussion}
This paper asked whether the information contained in the LLM activations, in the form of internal concept vectors, can be used to measure concepts from texts, and how it compares to the model's own response.
For the presence of concepts, it can: the linear probe outscores the model's own response in all twelve comparisons, indicating that the activations contain information that the model does not show in its response, and a median of 12.0 minutes for RFM and 43.4 for the probe per model and pillar, on 3 to 5 consumer GPUs (Appendix \ref{app:impl}), brings a frozen, out-of-the-box model close to the fine-tuned domain classifiers.
The surprise is which method performed the best.
We expected the RFM concept vectors of \citet{doi:10.1126/science.aea6792} to be the strongest extractor. Instead, the simpler linear probe was the best activation-based method, in all twelve accuracy comparisons on the same sentences \citep[cf.][]{pmlr-v267-wu25a}, and neither activation-based method overtook the fine-tuned domain classifiers of \citet{schimanski2024bridging}.

The concept vectors trailed the probe on every pillar, and on Environmental and Social fell below the embedding baseline on accuracy.
Larger models perform better than smaller ones on Environmental with the mean of five and on Social with the first concept vector, which is consistent with the findings of \citet{doi:10.1126/science.aea6792}.
The first concept vector also outperformed the mean of the top five concept vectors on accuracy in nine of the twelve model-pillar combinations, and on F1 in eight of twelve once the threshold is frozen on validation.
That second result is counterintuitive, as averaging several directions should capture more of a concept that is spread across the activation space.
A possible explanation is that the first concept vector, being the top eigenvector of the AGOP matrix, already captures the concept, while the additional vectors capture more peripheral concepts, which adds noise.
This explanation does not account for the size of the gap, which ranges from 0.002 to 0.052 in accuracy without tracking how difficult each pillar is.
Testing it directly would require inspecting what the second to fifth eigenvectors separate, which we leave to future work.

The two methods we used use the internal activations in two different ways.
The linear probe needs to find only a single boundary vector between the activations per layer that best separates them on their binary labels, and it does so reliably. A concept vector, however, must rely on the concept being present as a direction strong enough inside the model to be captured as a concept vector.
That is the harder task, as it is more prone to noise and isolating the exact concept you want inside the model is difficult, but this method is also the one that, in \citet{doi:10.1126/science.aea6792}, supports steering.
For measuring one concept with a labelled dataset being available \citep[e.g.,][]{zohrehvand2024event}, the probe is the better tool. The concept vectors' case rests on what they can offer beyond classification.

Appendix \ref{app:distributions} discusses the distributions of the RFM scores.
Appendix B discusses confident errors on sentences whose label is itself arguable.

Both extractors proved sensitive to the wrapper.
Ending it with ``Reason step-by-step. Final answer:'' rather than ``Answer:'' moves accuracy by at most 2.2 points on the 8B models and by 5.6 points for the concept vectors on Gemma-4-31B-it, and ``address Governance'' rather than ``discuss governance'' by at most 2.8, with the paired interval including zero in fifteen of the eighteen accuracy contrasts, while the seven wrappers we tried span 7 to 9 accuracy points for the concept vectors and 3 to 5 for the last-token probe (Appendix A).
Wrapper selection therefore was as important for the concept vectors as any other step we controlled, and the large gaps separate the wrappers that define the concept (W3 to W7) from the bare sentence and the short question (W1, W2), most likely because the definition helps steer the model towards isolating the concept.

Practical implications for measurement work are in Appendix~\ref{app:practical}.

\section{Conclusion}
This study asked whether the information contained in the LLM activations, in the form of internal concept vectors, can be used to measure concepts from texts, and how it compares to the model's own response.
We showed that it can, and does so cheaply.
First, the information contained in the activations outscores the model's own response as a source of measurement for the presence of concepts in all twelve comparisons, indicating that the activations contain information that the model does not show in its response.
Second, the best linear probe is 0.6 to 3.3 percentage points below the fine-tuned domain classifiers and needs a median of 43.4 minutes per model and pillar on 3 to 5 consumer GPUs, so leveraging the internal representation of an LLM can be a cost-effective and generalizable alternative to fine-tuning a domain classifier.
Out of the two methods tested there is a clear winner: the simpler linear probe consistently outperforms the RFM concept vectors.
What remains open is measuring the degree of concept presence in a text.
Both extractors yield a continuous score intended to reflect how much of a text concerns a concept, and a synthetic mixture test gives us first evidence that it does. However, current ESG datasets carry only binary labels, so validating the measurement of degree against human-graded labels remains open.
Future work should also develop a more structured method for designing the wrapper, which moves results by several points and remains the least controlled step in the pipeline.

\section*{Limitations}
The methods are sensitive to the wrapper, and our sweep covered seven wrappers on one pillar and three models, so how much better either method could do under an optimal wrapper is unknown.
This uncertainty has more impact on the RFM method than on the probe. This is because a concept vector must carry the concept, so a weak wrapper punishes it twice: once during training, when weak activations do not carry the concept and introduce noise, and once in evaluation, when weak concept vectors have lower similarity to new activations.
It therefore remains difficult to separate how much of RFM's weaker performance reflects the method and how much reflects our wrapper search.
The probe's wrapper was chosen partly on cross-validation accuracy over the same 2,000 sentences on which the CV probe is reported, so those numbers carry a selection optimism. The wrapper sweep in Appendix A bounds it, since across seven wrappers the probe's accuracy spans 3 to 5 points with last-token pooling, and the 8B models' chosen wrapper is the best of them under that pooling. With max pooling, Qwen-3-8B's span is 1.4 points and another wrapper is 0.6 points higher.

In addition to that, ESG ratings are a niche subset within the financial sector.
Our small models have likely seen a limited amount of ESG corpus during training, which may directly result in models having a less well-defined direction that captures each of the ESG pillars.
We see support for this directly in the results: Environmental sentences consistently score higher across all methods than Governance sentences.
This is likely due to Environmental topics coming up more often in textual data than the specifics of governance structures.
Using larger models, or a financially fine-tuned model like the base models of \citet{schimanski2024bridging}, would allow stronger directions of the ESG pillars to be present.

The Gemma-4-31B-it probe rows come from runs on five consumer cards and appear in all tables that compare the four models. Gemma-4-31B-it was also the only model loaded with 8-bit quantisation, a confound in comparisons between models.
The probe's cross-validation and RFM each use a single seed, so their confidence intervals reflect sentence resampling, not seed variation.

Lastly, fully validating the continuous RFM score as a measure of how much a concept is present, rather than only whether it is, is not yet possible in our domain.
Current ESG datasets, including the one we use \citep{schimanski2024bridging}, have only binary labels, so we leave a direct validation of the score's magnitude to future work.

\bibliography{references}

\begin{thebibliography}{45}
\providecommand{\natexlab}[1]{#1}

\bibitem[{Alain and Bengio(2017)}]{DBLP:conf/iclr/AlainB17}
Guillaume Alain and Yoshua Bengio. 2017.
\newblock Understanding intermediate layers using linear classifier probes.
\newblock In \emph{{5th International Conference on Learning Representations,
  {ICLR} 2017, Toulon, France, April 24-26, 2017, Workshop Track Proceedings}}.
  OpenReview.net.

\bibitem[{Azaria and Mitchell(2023)}]{azaria-mitchell-2023-internal}
Amos Azaria and Tom Mitchell. 2023.
\newblock \href {https://doi.org/10.18653/v1/2023.findings-emnlp.68} {The
  internal state of an {LLM} knows when it{'}s lying}.
\newblock In \emph{Findings of the Association for Computational Linguistics:
  EMNLP 2023}, pages 967--976, Singapore. Association for Computational
  Linguistics.

\bibitem[{Beaglehole et~al.(2026{\natexlab{a}})Beaglehole, Holzm{\"u}ller,
  Radhakrishnan, and Belkin}]{beaglehole2026xrfm}
Daniel Beaglehole, David Holzm{\"u}ller, Adityanarayanan Radhakrishnan, and
  Mikhail Belkin. 2026{\natexlab{a}}.
\newblock {xRFM}: Accurate, scalable, and interpretable feature learning models
  for tabular data.
\newblock In \emph{{The Fourteenth International Conference on Learning
  Representations ({ICLR})}}.

\bibitem[{Beaglehole et~al.(2026{\natexlab{b}})Beaglehole, Radhakrishnan,
  Boix-Adser{\`a}, and Belkin}]{doi:10.1126/science.aea6792}
Daniel Beaglehole, Adityanarayanan Radhakrishnan, Enric Boix-Adser{\`a}, and
  Mikhail Belkin. 2026{\natexlab{b}}.
\newblock \href {https://doi.org/10.1126/science.aea6792} {Toward universal
  steering and monitoring of {AI} models}.
\newblock \emph{Science}, 391(6787):787--792.

\bibitem[{Belinkov(2022)}]{10.1162/coli_a_00422}
Yonatan Belinkov. 2022.
\newblock \href {https://doi.org/10.1162/coli_a_00422} {Probing classifiers:
  Promises, shortcomings, and advances}.
\newblock \emph{Computational Linguistics}, 48(1):207--219.

\bibitem[{Bingler et~al.(2022)Bingler, Kraus, Leippold, and
  Webersinke}]{bingler2022cheap}
Julia~Anna Bingler, Mathias Kraus, Markus Leippold, and Nicolas Webersinke.
  2022.
\newblock \href {https://doi.org/10.1016/j.frl.2022.102776} {Cheap talk and
  cherry-picking: What {ClimateBert} has to say on corporate climate risk
  disclosures}.
\newblock \emph{Finance Research Letters}, 47:102776.

\bibitem[{Blei et~al.(2003)Blei, Ng, and Jordan}]{blei2003latent}
David~M. Blei, Andrew~Y. Ng, and Michael~I. Jordan. 2003.
\newblock Latent {Dirichlet} allocation.
\newblock \emph{Journal of Machine Learning Research}, 3:993--1022.

\bibitem[{Burns et~al.(2023)Burns, Ye, Klein, and
  Steinhardt}]{burns2023discovering}
Collin Burns, Haotian Ye, Dan Klein, and Jacob Steinhardt. 2023.
\newblock Discovering latent knowledge in language models without supervision.
\newblock In \emph{{The Eleventh International Conference on Learning
  Representations ({ICLR})}}.

\bibitem[{Choudhury et~al.(2026)Choudhury, Vanneste, and
  Zohrehvand}]{ChoudhuryEtAl2026}
Prithwiraj Choudhury, Bart Vanneste, and Amirhossein Zohrehvand. 2026.
\newblock \href {https://doi.org/10.2139/ssrn.4945933} {The {Wade} test:
  Generative {AI} and a {CEO} bot}.
\newblock Harvard Business School Working Paper No. 25-008.

\bibitem[{Du et~al.(2025)Du, Zhao, Mao, Xing, and Cambria}]{DU2025102755}
Kelvin Du, Yazhi Zhao, Rui Mao, Frank Xing, and Erik Cambria. 2025.
\newblock \href {https://doi.org/10.1016/j.inffus.2024.102755} {Natural
  language processing in finance: A survey}.
\newblock \emph{Information Fusion}, 115:102755.

\bibitem[{{Gemma Team}(2026)}]{gemmateam2026gemma4}
{Gemma Team}. 2026.
\newblock \href {https://arxiv.org/abs/2607.02770} {{Gemma} 4 technical
  report}.
\newblock \emph{Preprint}, arXiv:2607.02770.

\bibitem[{Gentzkow et~al.(2019)Gentzkow, Kelly, and Taddy}]{gentzkow2019text}
Matthew Gentzkow, Bryan Kelly, and Matt Taddy. 2019.
\newblock \href {https://doi.org/10.1257/jel.20181020} {Text as data}.
\newblock \emph{Journal of Economic Literature}, 57(3):535--574.

\bibitem[{Gilardi et~al.(2023)Gilardi, Alizadeh, and
  Kubli}]{gilardi2023chatgpt}
Fabrizio Gilardi, Meysam Alizadeh, and Ma{\"e}l Kubli. 2023.
\newblock \href {https://doi.org/10.1073/pnas.2305016120} {{ChatGPT}
  outperforms crowd workers for text-annotation tasks}.
\newblock \emph{Proceedings of the National Academy of Sciences},
  120(30):e2305016120.

\bibitem[{Grattafiori et~al.(2024)}]{grattafiori2024llama3}
Aaron Grattafiori et~al. 2024.
\newblock \href {https://arxiv.org/abs/2407.21783} {The {Llama} 3 herd of
  models}.
\newblock \emph{Preprint}, arXiv:2407.21783.

\bibitem[{Gurnee et~al.(2026)Gurnee, Sofroniew, Pearce, Piotrowski, Kauvar,
  Chen, Soligo, Bogdan, Ong, Wang, Thompson, Abrahams, Kantamneni, Ameisen,
  Batson, and Lindsey}]{gurnee2026workspace}
Wes Gurnee, Nicholas Sofroniew, Adam Pearce, Mateusz Piotrowski, Isaac Kauvar,
  Runjin Chen, Anna Soligo, Paul Bogdan, Euan Ong, Rowan Wang, T.~Ben Thompson,
  David Abrahams, Subhash Kantamneni, Emmanuel Ameisen, Joshua Batson, and Jack
  Lindsey. 2026.
\newblock \href {https://transformer-circuits.pub/2026/workspace/index.html}
  {Verbalizable representations form a global workspace in language models}.
\newblock \emph{Transformer Circuits Thread}.

\bibitem[{Gurnee and Tegmark(2024)}]{gurnee2024language}
Wes Gurnee and Max Tegmark. 2024.
\newblock Language models represent space and time.
\newblock In \emph{{The Twelfth International Conference on Learning
  Representations ({ICLR})}}.

\bibitem[{Hewitt and Liang(2019)}]{hewitt-liang-2019-designing}
John Hewitt and Percy Liang. 2019.
\newblock \href {https://doi.org/10.18653/v1/D19-1275} {Designing and
  interpreting probes with control tasks}.
\newblock In \emph{Proceedings of the 2019 Conference on Empirical Methods in
  Natural Language Processing and the 9th International Joint Conference on
  Natural Language Processing (EMNLP-IJCNLP)}, pages 2733--2743, Hong Kong,
  China. Association for Computational Linguistics.

\bibitem[{Hoerl and Kennard(1970)}]{Hoerl01021970}
Arthur~E. Hoerl and Robert~W. Kennard. 1970.
\newblock \href {https://doi.org/10.1080/00401706.1970.10488634} {Ridge
  regression: Biased estimation for nonorthogonal problems}.
\newblock \emph{Technometrics}, 12(1):55--67.

\bibitem[{Kim et~al.(2018)Kim, Wattenberg, Gilmer, Cai, Wexler, Vi{\'e}gas, and
  Sayres}]{pmlr-v80-kim18d}
Been Kim, Martin Wattenberg, Justin Gilmer, Carrie Cai, James Wexler, Fernanda
  Vi{\'e}gas, and Rory Sayres. 2018.
\newblock \href {https://proceedings.mlr.press/v80/kim18d.html}
  {Interpretability beyond feature attribution: Quantitative testing with
  concept activation vectors ({TCAV})}.
\newblock In \emph{Proceedings of the 35th International Conference on Machine
  Learning}, volume~80 of \emph{Proceedings of Machine Learning Research},
  pages 2668--2677. PMLR.

\bibitem[{Kozlowski et~al.(2019)Kozlowski, Taddy, and
  Evans}]{kozlowski2019geometry}
Austin~C. Kozlowski, Matt Taddy, and James~A. Evans. 2019.
\newblock \href {https://doi.org/10.1177/0003122419877135} {The geometry of
  culture: Analyzing the meanings of class through word embeddings}.
\newblock \emph{American Sociological Review}, 84(5):905--949.

\bibitem[{Li et~al.(2021)Li, Mai, Shen, and Yan}]{li2021measuring}
Kai Li, Feng Mai, Rui Shen, and Xinyan Yan. 2021.
\newblock \href {https://doi.org/10.1093/rfs/hhaa079} {Measuring corporate
  culture using machine learning}.
\newblock \emph{The Review of Financial Studies}, 34(7):3265--3315.

\bibitem[{Li et~al.(2023)Li, Patel, Vi{\'e}gas, Pfister, and
  Wattenberg}]{li2023inference}
Kenneth Li, Oam Patel, Fernanda Vi{\'e}gas, Hanspeter Pfister, and Martin
  Wattenberg. 2023.
\newblock \href
  {https://proceedings.neurips.cc/paper_files/paper/2023/hash/81b8390039b7302c909cb769f8b6cd93-Abstract-Conference.html}
  {Inference-time intervention: Eliciting truthful answers from a language
  model}.
\newblock In \emph{Advances in Neural Information Processing Systems},
  volume~36.

\bibitem[{Liu et~al.(2023)Liu, Casper, Hadfield-Menell, and
  Andreas}]{liu-etal-2023-cognitive}
Kevin Liu, Stephen Casper, Dylan Hadfield-Menell, and Jacob Andreas. 2023.
\newblock \href {https://doi.org/10.18653/v1/2023.emnlp-main.291} {Cognitive
  dissonance: Why do language model outputs disagree with internal
  representations of truthfulness?}
\newblock In \emph{Proceedings of the 2023 Conference on Empirical Methods in
  Natural Language Processing}, pages 4791--4797, Singapore. Association for
  Computational Linguistics.

\bibitem[{Loughran and McDonald(2011)}]{loughran2011liability}
Tim Loughran and Bill McDonald. 2011.
\newblock \href {https://doi.org/10.1111/j.1540-6261.2010.01625.x} {When is a
  liability not a liability? {T}extual analysis, dictionaries, and {10-Ks}}.
\newblock \emph{The Journal of Finance}, 66(1):35--65.

\bibitem[{Marks and Tegmark(2024)}]{marks2024geometry}
Samuel Marks and Max Tegmark. 2024.
\newblock The geometry of truth: Emergent linear structure in large language
  model representations of true/false datasets.
\newblock In \emph{{Conference on Language Modeling ({COLM})}}.

\bibitem[{Orgad et~al.(2025)Orgad, Toker, Gekhman, Reichart, Szpektor, Kotek,
  and Belinkov}]{orgad2025llms}
Hadas Orgad, Michael Toker, Zorik Gekhman, Roi Reichart, Idan Szpektor, Hadas
  Kotek, and Yonatan Belinkov. 2025.
\newblock \href
  {https://proceedings.iclr.cc/paper_files/paper/2025/hash/a712d461e57201efe35d429a6f1731c1-Abstract-Conference.html}
  {{LLMs} know more than they show: On the intrinsic representation of {LLM}
  hallucinations}.
\newblock In \emph{{The Thirteenth International Conference on Learning
  Representations ({ICLR})}}.

\bibitem[{Park et~al.(2024)Park, Choe, and Veitch}]{pmlr-v235-park24c}
Kiho Park, Yo~Joong Choe, and Victor Veitch. 2024.
\newblock The linear representation hypothesis and the geometry of large
  language models.
\newblock In \emph{{Proceedings of the 41st International Conference on Machine
  Learning}}, volume 235 of \emph{{Proceedings of Machine Learning Research}},
  pages 39643--39666. PMLR.

\bibitem[{Pedregosa et~al.(2011)Pedregosa, Varoquaux, Gramfort, Michel,
  Thirion, Grisel, Blondel, Prettenhofer, Weiss, Dubourg, Vanderplas, Passos,
  Cournapeau, Brucher, Perrot, and Duchesnay}]{pedregosa2011scikit}
Fabian Pedregosa, Ga{\"e}l Varoquaux, Alexandre Gramfort, Vincent Michel,
  Bertrand Thirion, Olivier Grisel, Mathieu Blondel, Peter Prettenhofer, Ron
  Weiss, Vincent Dubourg, Jake Vanderplas, Alexandre Passos, David Cournapeau,
  Matthieu Brucher, Matthieu Perrot, and {\'E}douard Duchesnay. 2011.
\newblock \href {https://jmlr.org/papers/v12/pedregosa11a.html}
  {{Scikit-learn}: Machine learning in {P}ython}.
\newblock \emph{Journal of Machine Learning Research}, 12(85):2825--2830.

\bibitem[{Radhakrishnan et~al.(2024)Radhakrishnan, Beaglehole, Pandit, and
  Belkin}]{radhakrishnan2024mechanism}
Adityanarayanan Radhakrishnan, Daniel Beaglehole, Parthe Pandit, and Mikhail
  Belkin. 2024.
\newblock \href {https://doi.org/10.1126/science.adi5639} {Mechanism for
  feature learning in neural networks and backpropagation-free machine learning
  models}.
\newblock \emph{Science}, 383(6690):1461--1467.

\bibitem[{Rathje et~al.(2024)Rathje, Mirea, Sucholutsky, Marjieh, Robertson,
  and Van~Bavel}]{rathje2024gpt}
Steve Rathje, Dan-Mircea Mirea, Ilia Sucholutsky, Raja Marjieh, Claire~E.
  Robertson, and Jay~J. Van~Bavel. 2024.
\newblock \href {https://doi.org/10.1073/pnas.2308950121} {{GPT} is an
  effective tool for multilingual psychological text analysis}.
\newblock \emph{Proceedings of the National Academy of Sciences},
  121(34):e2308950121.

\bibitem[{Reimers and Gurevych(2019)}]{reimers-gurevych-2019-sentence}
Nils Reimers and Iryna Gurevych. 2019.
\newblock \href {https://doi.org/10.18653/v1/D19-1410} {Sentence-{BERT}:
  Sentence embeddings using {S}iamese {BERT}-networks}.
\newblock In \emph{Proceedings of the 2019 Conference on Empirical Methods in
  Natural Language Processing and the 9th International Joint Conference on
  Natural Language Processing (EMNLP-IJCNLP)}, pages 3982--3992, Hong Kong,
  China. Association for Computational Linguistics.

\bibitem[{Schimanski et~al.(2023)Schimanski, Bingler, Kraus, Hyslop, and
  Leippold}]{schimanski-etal-2023-climatebert}
Tobias Schimanski, Julia Bingler, Mathias Kraus, Camilla Hyslop, and Markus
  Leippold. 2023.
\newblock \href {https://doi.org/10.18653/v1/2023.emnlp-main.975}
  {{C}limate{BERT}-{N}et{Z}ero: Detecting and assessing net zero and reduction
  targets}.
\newblock In \emph{Proceedings of the 2023 Conference on Empirical Methods in
  Natural Language Processing}, pages 15745--15756, Singapore. Association for
  Computational Linguistics.

\bibitem[{Schimanski et~al.(2024)Schimanski, Reding, Reding, Bingler, Kraus,
  and Leippold}]{schimanski2024bridging}
Tobias Schimanski, Andrin Reding, Nico Reding, Julia Bingler, Mathias Kraus,
  and Markus Leippold. 2024.
\newblock \href {https://doi.org/10.1016/j.frl.2024.104979} {Bridging the gap
  in {ESG} measurement: Using {NLP} to quantify environmental, social, and
  governance communication}.
\newblock \emph{Finance Research Letters}, 61:104979.

\bibitem[{Stammbach et~al.(2023)Stammbach, Webersinke, Bingler, Kraus, and
  Leippold}]{stammbach-etal-2023-environmental}
Dominik Stammbach, Nicolas Webersinke, Julia Bingler, Mathias Kraus, and Markus
  Leippold. 2023.
\newblock \href {https://doi.org/10.18653/v1/2023.acl-short.91} {Environmental
  claim detection}.
\newblock In \emph{Proceedings of the 61st Annual Meeting of the Association
  for Computational Linguistics (Volume 2: Short Papers)}, pages 1051--1066,
  Toronto, Canada. Association for Computational Linguistics.

\bibitem[{Tenney et~al.(2019)Tenney, Das, and Pavlick}]{tenney-etal-2019-bert}
Ian Tenney, Dipanjan Das, and Ellie Pavlick. 2019.
\newblock \href {https://doi.org/10.18653/v1/P19-1452} {{BERT} rediscovers the
  classical {NLP} pipeline}.
\newblock In \emph{Proceedings of the 57th Annual Meeting of the Association
  for Computational Linguistics}, pages 4593--4601, Florence, Italy.
  Association for Computational Linguistics.

\bibitem[{Vaswani et~al.(2017)Vaswani, Shazeer, Parmar, Uszkoreit, Jones,
  Gomez, Kaiser, and Polosukhin}]{NIPS2017_3f5ee243}
Ashish Vaswani, Noam Shazeer, Niki Parmar, Jakob Uszkoreit, Llion Jones,
  Aidan~N. Gomez, {\L}ukasz Kaiser, and Illia Polosukhin. 2017.
\newblock Attention is all you need.
\newblock In \emph{{Advances in Neural Information Processing Systems}},
  volume~30. Curran Associates, Inc.

\bibitem[{Webersinke et~al.(2021)Webersinke, Kraus, Bingler, and
  Leippold}]{webersinke2021climatebert}
Nicolas Webersinke, Mathias Kraus, Julia~Anna Bingler, and Markus Leippold.
  2021.
\newblock \href {https://arxiv.org/abs/2110.12010} {{ClimateBert}: A pretrained
  language model for climate-related text}.
\newblock \emph{Preprint}, arXiv:2110.12010.

\bibitem[{Wolf et~al.(2020)Wolf, Debut, Sanh, Chaumond, Delangue, Moi, Cistac,
  Rault, Louf, Funtowicz, Davison, Shleifer, von Platen, Ma, Jernite, Plu, Xu,
  Le~Scao, Gugger, Drame, Lhoest, and Rush}]{wolf-etal-2020-transformers}
Thomas Wolf, Lysandre Debut, Victor Sanh, Julien Chaumond, Clement Delangue,
  Anthony Moi, Pierric Cistac, Tim Rault, Remi Louf, Morgan Funtowicz, Joe
  Davison, Sam Shleifer, Patrick von Platen, Clara Ma, Yacine Jernite, Julien
  Plu, Canwen Xu, Teven Le~Scao, Sylvain Gugger, and 3 others. 2020.
\newblock \href {https://doi.org/10.18653/v1/2020.emnlp-demos.6} {Transformers:
  State-of-the-art natural language processing}.
\newblock In \emph{Proceedings of the 2020 Conference on Empirical Methods in
  Natural Language Processing: System Demonstrations}, pages 38--45, Online.
  Association for Computational Linguistics.

\bibitem[{Wu et~al.(2025)Wu, Arora, Geiger, Wang, Huang, Jurafsky, Manning, and
  Potts}]{pmlr-v267-wu25a}
Zhengxuan Wu, Aryaman Arora, Atticus Geiger, Zheng Wang, Jing Huang, Dan
  Jurafsky, Christopher~D. Manning, and Christopher Potts. 2025.
\newblock \href {https://proceedings.mlr.press/v267/wu25a.html} {{A}x{B}ench:
  Steering {LLM}s? {E}ven simple baselines outperform sparse autoencoders}.
\newblock In \emph{Proceedings of the 42nd International Conference on Machine
  Learning}, volume 267 of \emph{Proceedings of Machine Learning Research},
  pages 67035--67080. PMLR.

\bibitem[{Yang et~al.(2025)Yang, Li, Yang, Zhang, Hui, Zheng, Yu, Gao, Huang,
  Lv, Zheng, Liu, Zhou, Huang, Hu, Ge, Wei, Lin, Tang, Yang, Tu, Zhang, Yang,
  Yang, Zhou, Zhou, Lin, Dang, Bao, Yang, Yu, Deng, Li, Xue, Li, Zhang, Wang,
  Zhu, Men, Gao, Liu, Luo, Li, Tang, Yin, Ren, Wang, Zhang, Ren, Fan, Su,
  Zhang, Zhang, Wan, Liu, Wang, Cui, Zhang, Zhou, and Qiu}]{yang2025qwen3}
An~Yang, Anfeng Li, Baosong Yang, Beichen Zhang, Binyuan Hui, Bo~Zheng, Bowen
  Yu, Chang Gao, Chengen Huang, Chenxu Lv, Chujie Zheng, Dayiheng Liu, Fan
  Zhou, Fei Huang, Feng Hu, Hao Ge, Haoran Wei, Huan Lin, Jialong Tang, and 41
  others. 2025.
\newblock \href {https://arxiv.org/abs/2505.09388} {{Qwen3} technical report}.
\newblock \emph{Preprint}, arXiv:2505.09388.

\bibitem[{Yang et~al.(2024)Yang, Duan, Liu, and Tam}]{yang2024llmmeasure}
Yi~Yang, Hanyu Duan, Jiaxin Liu, and Kar~Yan Tam. 2024.
\newblock \href {https://arxiv.org/abs/2409.12722} {{LLM-Measure}: Generating
  valid, consistent, and reproducible text-based measures for social science
  research}.
\newblock \emph{Preprint}, arXiv:2409.12722.

\bibitem[{Zhang et~al.(2025)Zhang, Li, Long, Zhang, Lin, Yang, Xie, Yang, Liu,
  Lin, Huang, and Zhou}]{zhang2025qwen3embedding}
Yanzhao Zhang, Mingxin Li, Dingkun Long, Xin Zhang, Huan Lin, Baosong Yang,
  Pengjun Xie, An~Yang, Dayiheng Liu, Junyang Lin, Fei Huang, and Jingren Zhou.
  2025.
\newblock \href {https://arxiv.org/abs/2506.05176} {{Qwen3} embedding:
  Advancing text embedding and reranking through foundation models}.
\newblock \emph{Preprint}, arXiv:2506.05176.

\bibitem[{Ziems et~al.(2024)Ziems, Held, Shaikh, Chen, Zhang, and
  Yang}]{ziems2024can}
Caleb Ziems, William Held, Omar Shaikh, Jiaao Chen, Zhehao Zhang, and Diyi
  Yang. 2024.
\newblock \href {https://doi.org/10.1162/coli_a_00502} {Can large language
  models transform computational social science?}
\newblock \emph{Computational Linguistics}, 50(1):237--291.

\bibitem[{Zohrehvand et~al.(2024)Zohrehvand, Doshi, and
  Vanneste}]{zohrehvand2024event}
Amirhossein Zohrehvand, Anil~R. Doshi, and Bart~S. Vanneste. 2024.
\newblock \href {https://doi.org/10.1016/j.lrp.2023.102392} {Generalizing event
  studies using synthetic controls: An application to the {Dollar
  Tree}--{Family Dollar} acquisition}.
\newblock \emph{Long Range Planning}, 57(1):102392.

\bibitem[{Zou et~al.(2023)Zou, Phan, Chen, Campbell, Guo, Ren, Pan, Yin,
  Mazeika, Dombrowski, Goel, Li, Byun, Wang, Mallen, Basart, Koyejo, Song,
  Fredrikson, Kolter, and Hendrycks}]{zou2023transparency}
Andy Zou, Long Phan, Sarah Chen, James Campbell, Phillip Guo, Richard Ren,
  Alexander Pan, Xuwang Yin, Mantas Mazeika, Ann-Kathrin Dombrowski, Shashwat
  Goel, Nathaniel Li, Michael~J. Byun, Zifan Wang, Alex Mallen, Steven Basart,
  Sanmi Koyejo, Dawn Song, Matt Fredrikson, and 2 others. 2023.
\newblock \href {https://arxiv.org/abs/2310.01405} {Representation engineering:
  A top-down approach to {AI} transparency}.
\newblock \emph{Preprint}, arXiv:2310.01405.

\end{thebibliography}

\appendix

\section{Wrapper sensitivity}
\label{app:wrappers}
Table \ref{tab:wrapper-sensitivity} reports test accuracy on the Governance pillar for seven wrappers (W1: the bare sentence, W2: a short question, W3 to W7: variants of the definitional wrapper) with Qwen-3-8B and Llama-3.1-8B, each method trained on the same 1,000 training sentences and scored on the same 500 test sentences, RFM with thresholds frozen on validation and the probe with last-token pooling.
The sweep ran in the independent reimplementation (Appendix \ref{app:impl}), not the main pipeline. W7 and W6 are the wrapper texts of the main Governance runs of these two models for RFM and the probe, and the sweep's W7 RFM cells differ from Table \ref{tab:main} by at most 0.8 points, its W6 probe cells from the split probe of Table \ref{tab:full_governance} by 0.2 (Llama-3.1-8B) and 1.8 points (Qwen-3-8B).
Across the seven wrappers, the RFM first vector spans 9.0 accuracy points on Qwen-3-8B and 8.6 on Llama-3.1-8B, the mean of five 8.2 and 7.0, the last-token probe 3.2 and 4.0 and the mean-pooled probe 0.6 and 2.4. On Gemma-4-31B-it (Table \ref{tab:wrapper_sweep_gemma}) the pattern is the same: the RFM first vector spans 7.6 accuracy points across the seven wrappers (0.780 to 0.856) and the mean of five 9.0 points (0.756 to 0.846), the last-token probe 4.8 points (0.830 to 0.878) and the mean-pooled probe 1.6 points (0.854 to 0.870). On every wrapper the last-token probe beats both RFM variants (7 of 7 on accuracy and on AUC).

\begin{table*}[tp]
  \centering
  \footnotesize
  \setlength{\tabcolsep}{4pt}
  \begin{tabular}{@{}l@{\hspace{6pt}}c@{\hspace{6pt}}c@{\hspace{6pt}}c@{\hspace{10pt}}c@{\hspace{6pt}}c@{\hspace{6pt}}c@{}}
\toprule
 & \multicolumn{3}{c}{Qwen-3-8B} & \multicolumn{3}{c}{Llama-3.1-8B} \\
\cmidrule(lr){2-4} \cmidrule(lr){5-7}
Wrapper & RFM first & RFM mean of 5 & Probe & RFM first & RFM mean of 5 & Probe \\
\midrule
W1 & 0.764 {\scriptsize[0.726, 0.800]} & 0.752 {\scriptsize[0.714, 0.788]} & 0.826 {\scriptsize[0.792, 0.858]} & 0.758 {\scriptsize[0.720, 0.794]} & 0.766 {\scriptsize[0.728, 0.802]} & 0.834 {\scriptsize[0.800, 0.866]} \\
W2 & 0.776 {\scriptsize[0.740, 0.812]} & 0.782 {\scriptsize[0.746, 0.818]} & 0.832 {\scriptsize[0.800, 0.864]} & 0.750 {\scriptsize[0.712, 0.788]} & 0.798 {\scriptsize[0.762, 0.832]} & 0.844 {\scriptsize[0.812, 0.876]} \\
W3 & 0.814 {\scriptsize[0.780, 0.848]} & 0.804 {\scriptsize[0.770, 0.838]} & 0.846 {\scriptsize[0.814, 0.878]} & 0.794 {\scriptsize[0.758, 0.828]} & \textbf{0.836} {\scriptsize[0.802, 0.868]} & 0.868 {\scriptsize[0.838, 0.898]} \\
W4 & 0.848 {\scriptsize[0.816, 0.880]} & \textbf{0.834} {\scriptsize[0.802, 0.866]} & 0.838 {\scriptsize[0.806, 0.870]} & 0.814 {\scriptsize[0.780, 0.848]} & 0.834 {\scriptsize[0.802, 0.866]} & 0.870 {\scriptsize[0.840, 0.898]} \\
W5 & 0.850 {\scriptsize[0.818, 0.880]} & 0.800 {\scriptsize[0.764, 0.834]} & 0.852 {\scriptsize[0.820, 0.882]} & 0.828 {\scriptsize[0.794, 0.860]} & 0.830 {\scriptsize[0.796, 0.862]} & 0.866 {\scriptsize[0.836, 0.896]} \\
W6 & \textbf{0.854} {\scriptsize[0.822, 0.884]} & 0.828 {\scriptsize[0.796, 0.860]} & \textbf{0.858} {\scriptsize[0.828, 0.888]} & 0.814 {\scriptsize[0.780, 0.846]} & 0.830 {\scriptsize[0.796, 0.862]} & \textbf{0.874} {\scriptsize[0.844, 0.902]} \\
W7 & 0.832 {\scriptsize[0.800, 0.864]} & 0.806 {\scriptsize[0.770, 0.840]} & 0.854 {\scriptsize[0.824, 0.884]} & \textbf{0.836} {\scriptsize[0.802, 0.868]} & \textbf{0.836} {\scriptsize[0.802, 0.868]} & 0.864 {\scriptsize[0.834, 0.894]} \\
\bottomrule
\end{tabular}

  \vspace{6pt}
  \begin{tabular}{@{}l>{\raggedright\arraybackslash}p{14.3cm}@{}}
  \toprule
  Wrapper & Exact text \\
  \midrule
  W1 & \{statement\} \\
  W2 & Does the following sentence discuss corporate governance? Sentence: \{statement\} Final answer:\textvisiblespace{} \\
  W3 & Text: \{statement\} Does this text discuss Governance as accountability or oversight \textemdash{} such as board structures, ethics codes, anti-corruption, transparency in decision-making, or regulatory compliance? Let's think step by step. Final answer: \\
  W5 & Text: \{statement\} Does this text address Governance as accountability or oversight \textemdash{} such as board structures, ethics codes, anti-corruption, transparency in decision-making, or regulatory compliance? Reason step-by-step. Final answer: \\
  \bottomrule
  \end{tabular}
  \caption{Wrapper sensitivity on Governance. Test accuracy (500 sentences) with 95\% bootstrap CIs (10,000 resamples, seed 0) for the RFM first concept vector, the RFM mean of the top five vectors (threshold fitted on validation and applied frozen to test), and the stacked linear probe (last-token pooling, fixed 0.5 threshold), all trained on the same 1,000 training sentences. W6 and W7 have the same wrapper texts as the main Governance runs of Qwen-3-8B and Llama-3.1-8B (W6 for the probe, W7 for RFM). The numbers here come from the independent reimplementation (Appendix \ref{app:impl}). Exact strings of W1, W2, W3 and W5 below the table, with \{statement\} the slot for the sentence and \textvisiblespace{} a trailing space. W4, W6 and W7 appear verbatim in Table \ref{tab:wrappers_used} as \texttt{27deb551}, \texttt{6a5adb9f} and \texttt{a6d40e71}. Bold: best wrapper per column.}
  \label{tab:wrapper-sensitivity}
\end{table*}

\begin{table*}[tp]
  \centering
  \footnotesize
  \setlength{\tabcolsep}{4pt}
  \begin{tabular}{@{}l@{\hspace{6pt}}c@{\hspace{6pt}}c@{\hspace{6pt}}c@{}}
\toprule
 & \multicolumn{3}{c}{Gemma-4-31B-it (8-bit)} \\
\cmidrule(lr){2-4}
Wrapper & RFM first & RFM mean of 5 & Probe \\
\midrule
W1 & 0.780 {\scriptsize[0.744, 0.816]} & 0.756 {\scriptsize[0.718, 0.794]} & 0.830 {\scriptsize[0.796, 0.862]} \\
W2 & 0.808 {\scriptsize[0.772, 0.842]} & 0.804 {\scriptsize[0.770, 0.838]} & 0.866 {\scriptsize[0.836, 0.896]} \\
W3 & 0.840 {\scriptsize[0.808, 0.872]} & 0.828 {\scriptsize[0.794, 0.860]} & 0.862 {\scriptsize[0.832, 0.892]} \\
W4 & 0.784 {\scriptsize[0.748, 0.820]} & 0.842 {\scriptsize[0.810, 0.874]} & 0.862 {\scriptsize[0.832, 0.892]} \\
W5 & 0.820 {\scriptsize[0.786, 0.854]} & 0.762 {\scriptsize[0.724, 0.798]} & \textbf{0.878} {\scriptsize[0.848, 0.906]} \\
W6 & 0.800 {\scriptsize[0.764, 0.834]} & 0.790 {\scriptsize[0.754, 0.826]} & 0.876 {\scriptsize[0.848, 0.904]} \\
W7 & \textbf{0.856} {\scriptsize[0.824, 0.886]} & \textbf{0.846} {\scriptsize[0.814, 0.876]} & 0.870 {\scriptsize[0.840, 0.898]} \\
\bottomrule
\end{tabular}

  \caption{Wrapper sensitivity on Governance for Gemma-4-31B-it, loaded in 8-bit (bitsandbytes LLM.int8, as in the main runs) on one H100 GPU of a national HPC cluster. The 8B models in Table \ref{tab:wrapper-sensitivity} ran unquantised in bfloat16. Test accuracy (500 sentences) with 95\% bootstrap CIs (10,000 resamples, seed 0) for the RFM first concept vector, the RFM mean of the top five vectors (threshold fitted on validation and applied frozen to test), and the stacked linear probe (last-token pooling, fixed 0.5 threshold), all trained on the same 1,000 training sentences. Bold: best wrapper per column.}
  \label{tab:wrapper_sweep_gemma}
\end{table*}

The wrapper is also not model-agnostic: the same wrapper applied to different models gives different results relative to each model's performance.
For the first concept vector, the best wrapper is W6 on Qwen-3-8B and W7 on Llama-3.1-8B, and for the probe W6 on both, though the best wrappers lie within each other's intervals.
A possible explanation is that this depends on what each model has already learned.
Some models may have seen more financial ESG data during training than others, so a wrapper that explains the concept in more detail may help guide a model that knows less about the topic, while adding noise to one that already knows it well.

\section{Confidently misclassified RFM continuous scores}
The RFM method provides us with a score of how similar new activations are compared to the concept vector.
We scale the results between -1 and 1 in order to have a clear continuous score. This score can be interpreted as how much a concept is active in a new sentence and how confident the model is.
Analysing these results shows that the model sometimes very confidently misclassifies sentences from the ESG dataset created with three human annotators by \citet{schimanski2024bridging}.
However, the inter-annotator agreement is never 100\% on the datasets, and the lowest agreement is on the dataset that has proven to be the most difficult to score accurately across all methods, the Governance dataset.
When we look at the continuous scores our RFM method provides, some of the confident errors fall on sentences whose label is itself arguable. Figure \ref{fig:score-distributions} shows the distribution of the scores, and Table \ref{tab:misclassified} shows the top 5 most confidently misclassified sentences on Governance. Here we can see that some of the false positive sentences do mention governance topics, and that these topics are absent in some false negative cases. We see a similar case on the Environmental task: using Gemma-4-31B-it, the method scored this sentence in the high range of the positives: ``We are committed to creating a successful, sustainable business that is economically, socially and environmentally conscious.'' However, the dataset labels it negative. A new team of domain experts should re-annotate these sentences in order to make real claims on whether the label or the method is wrong.

\begin{table*}[tp]
  \centering
  \begin{tabular}{cp{11cm}}
\toprule
Score & Sentence \\
\midrule
\multicolumn{2}{l}{\textit{False positives (Pred = 1, Label = 0)}} \\
\addlinespace
$+0.869$ & Yes In the absence of business ties between Mr Jean Charest and Tikehau Capital, as the service agreement between the Company and Chardi Inc, a company of which Jean Charest is manager and shareholder, ended on 31~December 2018, the Supervisory Board considered that all the criteria set out in Article~8.5 of the AFEP-MEDEF Code were met. \\
$+0.844$ & If we fail to satisfy regulatory requirements or maintain our financial holding company status, our financial condition and results of operations could be adversely affected, and we may be restricted in our ability to take certain capital actions (such as declaring dividends or repurchasing outstanding shares) or engage in certain business activities or acquisitions, which could compromise our competitive position. \\
$+0.837$ & These bodies ensure that a disciplined risk culture is promoted, and thus risk transparency is enhanced. \\
$+0.814$ & Examples of such other laws and regulations are those relating to anti-bribery and corruption, competition and data privacy laws, and human rights. \\
$+0.804$ & In addition, each director has a statutory duty to disclose actual or potential conflicts of interest. \\
\addlinespace
\midrule
\multicolumn{2}{l}{\textit{False negatives (Pred = 0, Label = 1)}} \\
\addlinespace
$-0.717$ & We are building an inclusive culture, ensuring our people feel a strong sense of belonging, providing fair and progressive policies, equal pay for equal work, listening and dialogue sessions. \\
$-0.611$ & Relevance to strategy Human capital is important to online gaming businesses, and online businesses generally, and competition for highly-qualified personnel is intense in locations in which the Group is based. \\
$-0.466$ & \textbullet{} Accounting policies When the firm obtains control over an entity, the initial accounting for its assets and liabilities is at fair value. \\
$-0.463$ & HUMAN CAPITAL MANAGEMENT The Company strives to create a culture that not only drives strategic and key business priorities forward, but is welcoming, inclusive, diverse and encourages associates to create a positive impact in their global communities. \\
$-0.436$ & We use a combination of direct, indirect and strategic engagement approaches to influence the actions of governments and others. \\
\bottomrule
\end{tabular}

  \caption{The five most confident errors of RFM (mean of the top 5 concept vectors, Gemma-4-31B-it) on the Governance test split: the highest-scoring sentences labelled 0 and the lowest-scoring sentences labelled 1. Score: cosine with the concept vector, min-max scaled to [-1, 1] on the test split. Predictions use the threshold fitted on the validation split and frozen (0.385 on this scale). The test split has 59 false positives and 36 false negatives in all.}
  \label{tab:misclassified}
\end{table*}

\section{Implementation details}
\label{app:impl}
Llama-3.1-8B is a decoder-only model consisting of 32 layers containing a total of 8 billion parameters.
Previous research has already shown that this model performs well on activation-based tasks \citep{doi:10.1126/science.aea6792}.

From the Qwen-3 family we use two models, Qwen-3-8B and Qwen-3-14B.
Both models are decoder-only models, the 8b model consisting of 36 layers containing a total of 8 billion parameters and the 14b model consisting of 40 layers and 14 billion parameters.
These models were released in 2025 and have been demonstrated to perform well on reasoning tasks.
We expect that these models should be better at understanding more complex questions.

Gemma-4-31B-it is a decoder-only model consisting of 60 layers containing a total of 31 billion parameters.
Gemma-4-31B-it was loaded with 8-bit quantisation (bitsandbytes, bfloat16 compute). Llama-3.1-8B, Qwen-3-8B and Qwen-3-14B were loaded in float16 without quantisation.
This model was released in 2026 and is the biggest and newest model we use.
Our expectations are that this will be the best model for complex tasks and concept vectors.

Table \ref{tab:runinfo} lists the four LLMs \citep{grattafiori2024llama3, yang2025qwen3, gemmateam2026gemma4} with their Hugging Face IDs, the number of layers whose activations are used, and the logged compute.
Table \ref{tab:wrappers_used} lists the wrapper behind every reported cell.
The embedding model is \texttt{Qwen/\allowbreak Qwen3-Embedding-8B} \citep{zhang2025qwen3embedding}.
Llama-3.1-8B is used under the Llama 3.1 Community License, and Qwen-3-8B, Qwen-3-14B, Gemma-4-31B-it and Qwen-3-Embedding-8B under Apache-2.0.
The data are the Hugging Face datasets \texttt{ESGBERT/\allowbreak environmental\_2k}, \texttt{ESGBERT/\allowbreak social\_2k} and \texttt{ESGBERT/\allowbreak governance\_2k} \citep{schimanski2024bridging}. Of these, \texttt{environmental\_2k} is released under Apache-2.0, and the other two have no dataset card and therefore no licence statement.

\begin{table}[t]
  \centering
  \small
  \begin{tabular}{@{}lcc@{}}
    \toprule
    Domain & Fleiss' $\kappa$ & 3-of-3 agreement \\

    \midrule
    Environmental & 0.906 & 89.8\% \\
    Social        & 0.926 & 92.3\% \\
    Governance    & 0.867 & 87.0\% \\

    \bottomrule
  \end{tabular}
  \caption{Inter-annotator agreement across the three ESG domains.}
  \label{tab:iaa}
\end{table}

\begin{table}[t]
  \centering
  \small
  \begin{tabular}{@{}lrrr@{}}
    \toprule
    Dataset & Yes (True) & No (False) & Total \\

    \midrule
    Environmental & 658 & 1{,}342 & 2{,}000 \\
    Social        & 804 & 1{,}196 & 2{,}000 \\
    Governance    & 538 & 1{,}462 & 2{,}000 \\

    \bottomrule
  \end{tabular}
  \caption{Label distribution in the 2k expert-annotated ESG datasets, counted from the released data.}
  \label{tab:label_dist}
\end{table}
The seeds are 23 for the dataset shuffle (training sentences 0 to 999, validation 1,000 to 1,499 and test 1,500 to 1,999 after shuffling), 0 for RFM, 992 for the probe cross-validation and 0 for the synthetic paragraphs.
For the own-answer baseline, the prompt is given as the user's turn in the model's chat template, followed by the start of the model's reply, with thinking disabled for the Qwen models.
In the final runs, only Gemma-4-31B-it on Governance had more than one probe wrapper. There the wrapper was chosen separately for each pooling, by the validation accuracy of the probe trained on the 1,000 training sentences.
The fitted logistic regression model also yields one coefficient per layer.
We do not read these as a measure of where the concept sits, because the layer scores entering the meta-learner are highly correlated and the resulting coefficients are unstable across folds.
Locating the signal by depth would require scoring each layer on its own.
The software versions are PyTorch 2.10.0 and Hugging Face Transformers 5.5.4 \citep{wolf-etal-2020-transformers}, and the probes and baselines use scikit-learn \citep{pedregosa2011scikit}.
The logged wall-clock time per model-pillar cell has a median of 12.0 minutes for RFM and 43.4 minutes for the probe, on 3 to 5 NVIDIA GeForce RTX 2080 Ti cards per job.
The surface baselines and the H100 timings come from an independent reimplementation (GPU runs on an H100 node, surface baselines on CPU). The wrapper sweep of Appendix A was run with the same reimplementation.
The Gemma-4-31B-it sweep ran on a national HPC cluster with one H100 per wrapper and 8-bit bitsandbytes quantisation, in about 33 minutes per wrapper for RFM and the three probe poolings. The four longest wrappers (W3 to W6) needed more than the 180 GB of host memory that comes with one GPU to hold the RFM activations and ran on exclusive nodes.

\begin{table*}[tp]
  \centering
  \small
  \begin{tabular}{@{}llrrrr@{}}
\toprule
 & & & & \multicolumn{2}{c}{Minutes, median} \\
\cmidrule(lr){5-6}
Model & Hugging Face id & Layers & GPUs & RFM & Probe \\
\midrule
Gemma-4-31B-it & \texttt{google/gemma-4-31B-it} & 59 & 5 & 56.3 & 93.0 \\
Llama-3.1-8B & \texttt{meta-llama/Llama-3.1-8B-Instruct} & 31 & 3 & 8.7 & 32.2 \\
Qwen-3-8B & \texttt{Qwen/Qwen3-8B} & 35 & 3 & 9.4 & 34.9 \\
Qwen-3-14B & \texttt{Qwen/Qwen3-14B} & 39 & 3 & 13.3 & 45.5 \\
\bottomrule
\end{tabular}

  \caption{Models and compute. Layers: number of layers whose activations are used. Minutes: wall-clock time per pillar as logged, median over the three pillars, for the RFM run and the probe run of the selected wrapper. GPUs: NVIDIA GeForce RTX 2080 Ti cards per job.}
  \label{tab:runinfo}
\end{table*}

\begin{table*}[tp]
  \centering
  \footnotesize
  \begin{tabular*}{\linewidth}{@{\extracolsep{\fill}}lcccc@{}}
\toprule
Model & RFM & Probe, primary pooling & Probe, last/mean/max & Own answer \\
\midrule
\multicolumn{5}{l}{\textit{Environmental}} \\
Gemma-4-31B-it & \texttt{c1681aa5} & \texttt{c1681aa5} (last) & same & \texttt{c1681aa5} \\
Llama-3.1-8B & \texttt{c1681aa5} & \texttt{50ec1cc0} (last) & same & \texttt{c1681aa5} \\
Qwen-3-8B & \texttt{c1681aa5} & \texttt{50ec1cc0} (last) & same & \texttt{c1681aa5} \\
Qwen-3-14B & \texttt{9d5e19d5} & \texttt{f878927c} (last) & same & \texttt{c1681aa5} \\
\midrule
\multicolumn{5}{l}{\textit{Social}} \\
Gemma-4-31B-it & \texttt{279ef1f9} & \texttt{279ef1f9} (mean) & same & \texttt{208123e6} \\
Llama-3.1-8B & \texttt{279ef1f9} & \texttt{1adfd543} (max) & same & \texttt{208123e6} \\
Qwen-3-8B & \texttt{279ef1f9} & \texttt{1adfd543} (max) & same & \texttt{208123e6} \\
Qwen-3-14B & \texttt{279ef1f9} & \texttt{1adfd543} (max) & same & \texttt{208123e6} \\
\midrule
\multicolumn{5}{l}{\textit{Governance}} \\
Gemma-4-31B-it & \texttt{27deb551} & \texttt{32dcd69f} (max) & \texttt{d595b72b}/\texttt{a6d40e71}/\texttt{32dcd69f} & \texttt{4b8ce80b} \\
Llama-3.1-8B & \texttt{a6d40e71} & \texttt{6a5adb9f} (last) & same & \texttt{4b8ce80b} \\
Qwen-3-8B & \texttt{a6d40e71} & \texttt{6a5adb9f} (max) & same & \texttt{4b8ce80b} \\
Qwen-3-14B & \texttt{a6d40e71} & \texttt{6a5adb9f} (max) & same & \texttt{4b8ce80b} \\
\bottomrule
\end{tabular*}

\vspace{6pt}
\ifdefined\wrapperidwidth\else\newlength{\wrapperidwidth}\fi
\settowidth{\wrapperidwidth}{\texttt{00000000}}
\begin{tabular}{@{}l>{\raggedright\arraybackslash}p{\dimexpr\linewidth-\wrapperidwidth-2\tabcolsep\relax}@{}}
\toprule
Id & Exact text \\
\midrule
\texttt{c1681aa5} & Does the following sentence contain ESG Environmental content?\textbackslash{}n\textbackslash{}nSentence: \{statement\}\textbackslash{}n\textbackslash{}nAnswer (yes/no):\textvisiblespace{} \\
\texttt{50ec1cc0} & Text: \{statement\} Does this text discuss environmental sustainability, climate impact, or environmental risk? Reason step-by-step. Final answer: \\
\texttt{9d5e19d5} & Does the following sentence contain ESG Environmental content, as opposed to social, governance, or general financial content?\textbackslash{}n\textbackslash{}nSentence: \{statement\}\textbackslash{}n\textbackslash{}nDoes it contain ESG Environmental content? Answer (yes/no):\textvisiblespace{} \\
\texttt{f878927c} & Does the following sentence contain ESG Environmental content?\textbackslash{}n\textbackslash{}nSentence: \{statement\}\textbackslash{}n\textbackslash{}nDoes it contain ESG Environmental content? Answer (yes/no):\textvisiblespace{} \\
\texttt{279ef1f9} & ESG Social content concerns a company's responsibilities toward people: employees, diversity, human rights, and communities. Consider how much ESG Social content the following sentence contains. Sentence:\textbackslash{}n\textbackslash{}n\{statement\}\textbackslash{}n\textbackslash{}nThe amount of ESG Social content is:\textvisiblespace{} \\
\texttt{208123e6} & ESG Social content concerns a company's responsibilities toward people: employees, diversity, human rights, and communities. Consider if ESG Social content is in the following sentence. Sentence:\textbackslash{}n\textbackslash{}n\{statement\}\textbackslash{}n\textbackslash{}nFinal answer (Yes/No):\textvisiblespace{} \\
\texttt{1adfd543} & Text: \{statement\} Does this text describe how the company treats or affects its workforce, customers, suppliers, or community \textemdash{} including matters like employee demographics, wages, benefits, health and safety, training, diversity, human rights, or community programs? Reason step-by-step. Final answer:\textvisiblespace{} \\
\texttt{27deb551} & Text: \{statement\} Does this text discuss Governance as accountability or oversight \textemdash{} such as board structures, ethics codes, anti-corruption,transparency in decision-making, or regulatory compliance? Reason step-by-step. Answer: \\
\texttt{d595b72b} & Corporate governance concerns how a company is directed, controlled and held accountable: board and committee structures, audit and financial reporting, internal controls, executive remuneration, shareholder rights, ethics codes, anti-corruption, and regulatory compliance. Text: \{statement\} Does this text discuss corporate governance? Answer (yes/no): \\
\texttt{a6d40e71} & Text: \{statement\} Does this text discuss governance as accountability or oversight \textemdash{} such as board structures, ethics codes, anti-corruption, transparency in decision-making, or regulatory compliance? Answer:\textvisiblespace{} \\
\texttt{32dcd69f} & Text: \{statement\} Does this text discuss governance as accountability or oversight \textemdash{} such as board and committee structures, audit and financial reporting, internal controls, executive remuneration, shareholder rights, ethics codes, anti-corruption, or regulatory compliance? Answer (yes/no): \\
\texttt{4b8ce80b} & Text: \{statement\}\textbackslash{}nDoes this text discuss Governance as accountability or oversight \textemdash{} such as board structures, ethics codes, anti-corruption,transparency in decision-making, or regulatory compliance? Answer (yes/no):\textvisiblespace{} \\
\texttt{6a5adb9f} & Text: \{statement\} Does this text discuss governance as accountability or oversight \textemdash{} such as board structures, ethics codes, anti-corruption, transparency in decision-making, or regulatory compliance? Reason step-by-step. Final answer: \\
\addlinespace
\multicolumn{2}{l}{\textit{Tried for a probe, not selected}} \\
\texttt{9ca602ec} & \{statement\} Was the previous sentence about ESG Goverance? Answer yer or no. \\
\bottomrule
\end{tabular}

  \caption{Prompt wrappers behind every reported cell. An id is the first 8 hex digits of the MD5 of the exact wrapper string. The strings are listed verbatim below the table, with \textbackslash{}n marking a newline, \textvisiblespace{} a trailing space, and \{statement\} the slot that receives the sentence. RFM and the primary probe use the same wrapper in 2 of the 12 cells (Gemma-4-31B-it Environmental, Gemma-4-31B-it Social). Probe poolings share one wrapper except for Gemma-4-31B-it on Governance. Own answer: the yes/no prompt recorded with each RFM run.}
  \label{tab:wrappers_used}
\end{table*}

\section{Full results per pillar}
\label{app:full}
Tables \ref{tab:full_environmental}, \ref{tab:full_social} and \ref{tab:full_governance} report every metric for every method, including all three poolings of the probe under both cross-validation and the RFM split.
For Environmental, Social and Governance, the primary poolings ($\dagger$) are last, mean and max for Gemma-4-31B-it, last, max and last for Llama-3.1-8B, and last, max and max for both Qwen models.

\begin{table*}[tp]
  \centering
  \footnotesize
  \begin{tabular}{@{}l@{\hspace{6pt}}c@{\hspace{6pt}}c@{\hspace{6pt}}c@{\hspace{6pt}}c@{\hspace{6pt}}c@{}}
\toprule
Method & AUC & Acc & F1 & Precision & Recall \\
\midrule
\multicolumn{6}{l}{\textit{Model's own answer (test split, $n = 500$)}} \\
Gemma-4-31B-it & 0.973 {\scriptsize [0.959, 0.985]} & 0.930 {\scriptsize [0.906, 0.952]} & 0.905 {\scriptsize [0.871, 0.935]} & 0.865 {\scriptsize [0.814, 0.912]} & 0.949 {\scriptsize [0.914, 0.978]} \\
Llama-3.1-8B & 0.969 {\scriptsize [0.955, 0.981]} & 0.864 {\scriptsize [0.834, 0.892]} & 0.833 {\scriptsize [0.792, 0.870]} & 0.735 {\scriptsize [0.676, 0.791]} & 0.960 {\scriptsize [0.929, 0.988]} \\
Qwen-3-8B & 0.980 {\scriptsize [0.969, 0.989]} & 0.920 {\scriptsize [0.896, 0.942]} & 0.892 {\scriptsize [0.856, 0.923]} & 0.851 {\scriptsize [0.799, 0.898]} & 0.938 {\scriptsize [0.898, 0.971]} \\
Qwen-3-14B & 0.978 {\scriptsize [0.967, 0.988]} & 0.916 {\scriptsize [0.890, 0.940]} & 0.887 {\scriptsize [0.852, 0.920]} & 0.842 {\scriptsize [0.789, 0.892]} & 0.938 {\scriptsize [0.899, 0.971]} \\
\midrule
\multicolumn{6}{l}{\textit{RFM, first concept vector, frozen threshold (test split, $n = 500$)}} \\
Gemma-4-31B-it & 0.975 {\scriptsize [0.961, 0.987]} & 0.936 {\scriptsize [0.914, 0.956]} & 0.912 {\scriptsize [0.880, 0.941]} & 0.883 {\scriptsize [0.835, 0.927]} & 0.943 {\scriptsize [0.907, 0.975]} \\
Llama-3.1-8B & 0.962 {\scriptsize [0.946, 0.976]} & 0.894 {\scriptsize [0.866, 0.920]} & 0.845 {\scriptsize [0.800, 0.884]} & 0.873 {\scriptsize [0.820, 0.922]} & 0.818 {\scriptsize [0.759, 0.874]} \\
Qwen-3-8B & 0.965 {\scriptsize [0.951, 0.978]} & 0.892 {\scriptsize [0.864, 0.918]} & 0.850 {\scriptsize [0.808, 0.888]} & 0.832 {\scriptsize [0.775, 0.883]} & 0.869 {\scriptsize [0.818, 0.917]} \\
Qwen-3-14B & 0.984 {\scriptsize [0.975, 0.991]} & 0.936 {\scriptsize [0.914, 0.956]} & 0.910 {\scriptsize [0.876, 0.939]} & 0.900 {\scriptsize [0.853, 0.942]} & 0.920 {\scriptsize [0.878, 0.958]} \\
\midrule
\multicolumn{6}{l}{\textit{RFM, mean of 5 concept vectors, frozen threshold (test split, $n = 500$)}} \\
Gemma-4-31B-it & 0.983 {\scriptsize [0.973, 0.992]} & 0.940 {\scriptsize [0.918, 0.960]} & 0.915 {\scriptsize [0.883, 0.944]} & 0.910 {\scriptsize [0.866, 0.949]} & 0.920 {\scriptsize [0.878, 0.958]} \\
Llama-3.1-8B & 0.943 {\scriptsize [0.922, 0.961]} & 0.872 {\scriptsize [0.842, 0.900]} & 0.814 {\scriptsize [0.766, 0.855]} & 0.833 {\scriptsize [0.776, 0.886]} & 0.795 {\scriptsize [0.735, 0.852]} \\
Qwen-3-8B & 0.912 {\scriptsize [0.887, 0.935]} & 0.844 {\scriptsize [0.810, 0.876]} & 0.755 {\scriptsize [0.698, 0.806]} & 0.845 {\scriptsize [0.782, 0.903]} & 0.682 {\scriptsize [0.612, 0.751]} \\
Qwen-3-14B & 0.980 {\scriptsize [0.970, 0.989]} & 0.938 {\scriptsize [0.916, 0.958]} & 0.913 {\scriptsize [0.880, 0.942]} & 0.905 {\scriptsize [0.860, 0.946]} & 0.920 {\scriptsize [0.878, 0.958]} \\
\midrule
\multicolumn{6}{l}{\textit{Linear probe, 5-fold cross-validation over all 2,000 sentences}} \\
Gemma-4-31B-it, last\rlap{$^{\dagger}$} & 0.984 {\scriptsize [0.979, 0.989]} & 0.949 {\scriptsize [0.939, 0.958]} & 0.923 {\scriptsize [0.907, 0.937]} & 0.912 {\scriptsize [0.890, 0.933]} & 0.933 {\scriptsize [0.914, 0.952]} \\
Gemma-4-31B-it, mean & 0.982 {\scriptsize [0.975, 0.988]} & 0.948 {\scriptsize [0.938, 0.957]} & 0.921 {\scriptsize [0.906, 0.936]} & 0.910 {\scriptsize [0.887, 0.930]} & 0.933 {\scriptsize [0.913, 0.951]} \\
Gemma-4-31B-it, max & 0.983 {\scriptsize [0.977, 0.988]} & 0.945 {\scriptsize [0.935, 0.955]} & 0.916 {\scriptsize [0.900, 0.931]} & 0.909 {\scriptsize [0.886, 0.930]} & 0.924 {\scriptsize [0.903, 0.944]} \\
\addlinespace[2pt]
Llama-3.1-8B, last\rlap{$^{\dagger}$} & 0.985 {\scriptsize [0.979, 0.989]} & 0.945 {\scriptsize [0.935, 0.954]} & 0.916 {\scriptsize [0.900, 0.931]} & 0.913 {\scriptsize [0.891, 0.933]} & 0.919 {\scriptsize [0.898, 0.940]} \\
Llama-3.1-8B, mean & 0.983 {\scriptsize [0.977, 0.988]} & 0.944 {\scriptsize [0.933, 0.954]} & 0.914 {\scriptsize [0.898, 0.930]} & 0.911 {\scriptsize [0.889, 0.932]} & 0.918 {\scriptsize [0.896, 0.938]} \\
Llama-3.1-8B, max & 0.983 {\scriptsize [0.977, 0.988]} & 0.947 {\scriptsize [0.937, 0.957]} & 0.920 {\scriptsize [0.905, 0.935]} & 0.911 {\scriptsize [0.889, 0.931]} & 0.930 {\scriptsize [0.910, 0.949]} \\
\addlinespace[2pt]
Qwen-3-8B, last\rlap{$^{\dagger}$} & 0.983 {\scriptsize [0.977, 0.988]} & 0.951 {\scriptsize [0.941, 0.960]} & 0.925 {\scriptsize [0.910, 0.939]} & 0.920 {\scriptsize [0.899, 0.941]} & 0.930 {\scriptsize [0.910, 0.949]} \\
Qwen-3-8B, mean & 0.982 {\scriptsize [0.976, 0.987]} & 0.942 {\scriptsize [0.932, 0.952]} & 0.912 {\scriptsize [0.895, 0.927]} & 0.914 {\scriptsize [0.892, 0.935]} & 0.909 {\scriptsize [0.886, 0.930]} \\
Qwen-3-8B, max & 0.981 {\scriptsize [0.974, 0.987]} & 0.950 {\scriptsize [0.940, 0.960]} & 0.925 {\scriptsize [0.910, 0.939]} & 0.914 {\scriptsize [0.892, 0.935]} & 0.936 {\scriptsize [0.917, 0.954]} \\
\addlinespace[2pt]
Qwen-3-14B, last\rlap{$^{\dagger}$} & 0.982 {\scriptsize [0.976, 0.987]} & 0.943 {\scriptsize [0.933, 0.953]} & 0.914 {\scriptsize [0.898, 0.929]} & 0.909 {\scriptsize [0.887, 0.930]} & 0.919 {\scriptsize [0.898, 0.940]} \\
Qwen-3-14B, mean & 0.980 {\scriptsize [0.973, 0.985]} & 0.940 {\scriptsize [0.930, 0.951]} & 0.909 {\scriptsize [0.893, 0.925]} & 0.904 {\scriptsize [0.881, 0.926]} & 0.915 {\scriptsize [0.893, 0.936]} \\
Qwen-3-14B, max & 0.982 {\scriptsize [0.975, 0.987]} & 0.945 {\scriptsize [0.935, 0.955]} & 0.918 {\scriptsize [0.902, 0.932]} & 0.906 {\scriptsize [0.883, 0.926]} & 0.930 {\scriptsize [0.910, 0.949]} \\
\midrule
\multicolumn{6}{l}{\textit{Linear probe, RFM split (train 1,000, test $n = 500$)}} \\
Gemma-4-31B-it, last\rlap{$^{\dagger}$} & 0.988 {\scriptsize [0.981, 0.994]} & 0.942 {\scriptsize [0.920, 0.962]} & 0.919 {\scriptsize [0.888, 0.947]} & 0.906 {\scriptsize [0.861, 0.948]} & 0.932 {\scriptsize [0.892, 0.966]} \\
Gemma-4-31B-it, mean & 0.986 {\scriptsize [0.977, 0.992]} & 0.940 {\scriptsize [0.918, 0.960]} & 0.917 {\scriptsize [0.885, 0.944]} & 0.897 {\scriptsize [0.851, 0.939]} & 0.938 {\scriptsize [0.901, 0.971]} \\
Gemma-4-31B-it, max & 0.987 {\scriptsize [0.978, 0.994]} & 0.948 {\scriptsize [0.928, 0.966]} & 0.928 {\scriptsize [0.897, 0.954]} & 0.903 {\scriptsize [0.857, 0.943]} & 0.955 {\scriptsize [0.921, 0.983]} \\
\addlinespace[2pt]
Llama-3.1-8B, last\rlap{$^{\dagger}$} & 0.983 {\scriptsize [0.972, 0.991]} & 0.938 {\scriptsize [0.916, 0.958]} & 0.913 {\scriptsize [0.880, 0.942]} & 0.901 {\scriptsize [0.854, 0.943]} & 0.926 {\scriptsize [0.886, 0.962]} \\
Llama-3.1-8B, mean & 0.983 {\scriptsize [0.973, 0.991]} & 0.938 {\scriptsize [0.916, 0.958]} & 0.913 {\scriptsize [0.880, 0.941]} & 0.905 {\scriptsize [0.860, 0.946]} & 0.920 {\scriptsize [0.879, 0.958]} \\
Llama-3.1-8B, max & 0.987 {\scriptsize [0.978, 0.993]} & 0.938 {\scriptsize [0.916, 0.958]} & 0.914 {\scriptsize [0.883, 0.942]} & 0.896 {\scriptsize [0.850, 0.938]} & 0.932 {\scriptsize [0.893, 0.966]} \\
\addlinespace[2pt]
Qwen-3-8B, last\rlap{$^{\dagger}$} & 0.985 {\scriptsize [0.975, 0.993]} & 0.952 {\scriptsize [0.932, 0.970]} & 0.933 {\scriptsize [0.904, 0.958]} & 0.922 {\scriptsize [0.881, 0.959]} & 0.943 {\scriptsize [0.907, 0.975]} \\
Qwen-3-8B, mean & 0.983 {\scriptsize [0.973, 0.992]} & 0.936 {\scriptsize [0.914, 0.956]} & 0.909 {\scriptsize [0.874, 0.939]} & 0.914 {\scriptsize [0.870, 0.953]} & 0.903 {\scriptsize [0.857, 0.945]} \\
Qwen-3-8B, max & 0.983 {\scriptsize [0.972, 0.992]} & 0.948 {\scriptsize [0.928, 0.966]} & 0.928 {\scriptsize [0.898, 0.954]} & 0.908 {\scriptsize [0.864, 0.947]} & 0.949 {\scriptsize [0.914, 0.978]} \\
\addlinespace[2pt]
Qwen-3-14B, last\rlap{$^{\dagger}$} & 0.981 {\scriptsize [0.970, 0.990]} & 0.942 {\scriptsize [0.920, 0.962]} & 0.919 {\scriptsize [0.887, 0.946]} & 0.906 {\scriptsize [0.862, 0.946]} & 0.932 {\scriptsize [0.893, 0.967]} \\
Qwen-3-14B, mean & 0.982 {\scriptsize [0.972, 0.990]} & 0.936 {\scriptsize [0.914, 0.956]} & 0.910 {\scriptsize [0.876, 0.939]} & 0.904 {\scriptsize [0.859, 0.945]} & 0.915 {\scriptsize [0.871, 0.953]} \\
Qwen-3-14B, max & 0.987 {\scriptsize [0.979, 0.994]} & 0.946 {\scriptsize [0.926, 0.966]} & 0.924 {\scriptsize [0.893, 0.951]} & 0.912 {\scriptsize [0.868, 0.950]} & 0.938 {\scriptsize [0.900, 0.971]} \\
\midrule
\multicolumn{6}{l}{\textit{Surface and embedding baselines, 5-fold CV over 2,000 (mean $\pm$ sd over folds)}} \\
Dictionary + LR & 0.880 {\scriptsize $\pm$0.007} & 0.771 {\scriptsize $\pm$0.008} & 0.475 {\scriptsize $\pm$0.021} & 0.958 {\scriptsize $\pm$0.037} & 0.316 {\scriptsize $\pm$0.015} \\
LDA + LR & 0.827 {\scriptsize $\pm$0.026} & 0.777 {\scriptsize $\pm$0.025} & 0.573 {\scriptsize $\pm$0.063} & 0.770 {\scriptsize $\pm$0.059} & 0.459 {\scriptsize $\pm$0.067} \\
TF-IDF + LR & 0.954 {\scriptsize $\pm$0.011} & 0.883 {\scriptsize $\pm$0.017} & 0.801 {\scriptsize $\pm$0.031} & 0.904 {\scriptsize $\pm$0.023} & 0.719 {\scriptsize $\pm$0.038} \\
Embedding + LR & 0.981 {\scriptsize $\pm$0.007} & 0.943 {\scriptsize $\pm$0.005} & 0.916 {\scriptsize $\pm$0.008} & 0.884 {\scriptsize $\pm$0.007} & 0.950 {\scriptsize $\pm$0.023} \\
\midrule
\multicolumn{6}{l}{\textit{Fine-tuned classifiers (Schimanski et al., 2024, mean $\pm$ sd as published)}} \\
EnvRoBERTa & & 0.957 {\scriptsize $\pm$0.010} & 0.932 {\scriptsize $\pm$0.014} & 0.933 {\scriptsize $\pm$0.040} & 0.933 {\scriptsize $\pm$0.031} \\
EnvDistilRoBERTa & & 0.950 {\scriptsize $\pm$0.011} & 0.924 {\scriptsize $\pm$0.017} & 0.906 {\scriptsize $\pm$0.033} & 0.942 {\scriptsize $\pm$0.018} \\
\bottomrule
\end{tabular}

  \caption{All methods on the Environmental pillar. Own answer, RFM (threshold fitted on validation and frozen) and the split probe are scored on the 500-sentence test split, and the CV probe by 5-fold cross-validation over 2,000 sentences. Brackets: 95\% bootstrap confidence intervals. $\dagger$: the model's primary pooling (highest validation accuracy). Surface baselines: independent reimplementation (Appendix \ref{app:impl}), same fold seed (992). Dictionary: concept-word share. LDA: 25 topics. TF-IDF: word 1- and 2-grams. Embedding + LR: Qwen-3-Embedding-8B features.}
  \label{tab:full_environmental}
\end{table*}

\begin{table*}[tp]
  \centering
  \footnotesize
  \begin{tabular}{@{}l@{\hspace{6pt}}c@{\hspace{6pt}}c@{\hspace{6pt}}c@{\hspace{6pt}}c@{\hspace{6pt}}c@{}}
\toprule
Method & AUC & Acc & F1 & Precision & Recall \\
\midrule
\multicolumn{6}{l}{\textit{Model's own answer (test split, $n = 500$)}} \\
Gemma-4-31B-it & 0.860 {\scriptsize [0.822, 0.896]} & 0.804 {\scriptsize [0.768, 0.838]} & 0.750 {\scriptsize [0.699, 0.796]} & 0.750 {\scriptsize [0.688, 0.811]} & 0.750 {\scriptsize [0.689, 0.810]} \\
Llama-3.1-8B & 0.892 {\scriptsize [0.863, 0.919]} & 0.822 {\scriptsize [0.788, 0.854]} & 0.756 {\scriptsize [0.704, 0.802]} & 0.817 {\scriptsize [0.758, 0.873]} & 0.704 {\scriptsize [0.638, 0.767]} \\
Qwen-3-8B & 0.937 {\scriptsize [0.916, 0.955]} & 0.836 {\scriptsize [0.804, 0.868]} & 0.773 {\scriptsize [0.724, 0.819]} & 0.843 {\scriptsize [0.786, 0.895]} & 0.714 {\scriptsize [0.650, 0.776]} \\
Qwen-3-14B & 0.893 {\scriptsize [0.864, 0.920]} & 0.810 {\scriptsize [0.774, 0.844]} & 0.756 {\scriptsize [0.705, 0.803]} & 0.762 {\scriptsize [0.699, 0.822]} & 0.750 {\scriptsize [0.688, 0.810]} \\
\midrule
\multicolumn{6}{l}{\textit{RFM, first concept vector, frozen threshold (test split, $n = 500$)}} \\
Gemma-4-31B-it & 0.943 {\scriptsize [0.922, 0.961]} & 0.868 {\scriptsize [0.838, 0.896]} & 0.832 {\scriptsize [0.789, 0.870]} & 0.832 {\scriptsize [0.776, 0.882]} & 0.832 {\scriptsize [0.777, 0.882]} \\
Llama-3.1-8B & 0.922 {\scriptsize [0.897, 0.944]} & 0.852 {\scriptsize [0.820, 0.882]} & 0.812 {\scriptsize [0.768, 0.852]} & 0.808 {\scriptsize [0.751, 0.861]} & 0.816 {\scriptsize [0.760, 0.869]} \\
Qwen-3-8B & 0.907 {\scriptsize [0.881, 0.932]} & 0.818 {\scriptsize [0.784, 0.852]} & 0.787 {\scriptsize [0.743, 0.829]} & 0.727 {\scriptsize [0.670, 0.784]} & 0.857 {\scriptsize [0.807, 0.905]} \\
Qwen-3-14B & 0.935 {\scriptsize [0.914, 0.954]} & 0.854 {\scriptsize [0.822, 0.884]} & 0.807 {\scriptsize [0.762, 0.849]} & 0.836 {\scriptsize [0.780, 0.888]} & 0.781 {\scriptsize [0.721, 0.837]} \\
\midrule
\multicolumn{6}{l}{\textit{RFM, mean of 5 concept vectors, frozen threshold (test split, $n = 500$)}} \\
Gemma-4-31B-it & 0.922 {\scriptsize [0.900, 0.943]} & 0.834 {\scriptsize [0.802, 0.866]} & 0.773 {\scriptsize [0.724, 0.818]} & 0.834 {\scriptsize [0.777, 0.888]} & 0.719 {\scriptsize [0.657, 0.782]} \\
Llama-3.1-8B & 0.902 {\scriptsize [0.874, 0.928]} & 0.824 {\scriptsize [0.790, 0.858]} & 0.774 {\scriptsize [0.726, 0.819]} & 0.778 {\scriptsize [0.718, 0.835]} & 0.770 {\scriptsize [0.710, 0.828]} \\
Qwen-3-8B & 0.841 {\scriptsize [0.804, 0.873]} & 0.768 {\scriptsize [0.730, 0.804]} & 0.741 {\scriptsize [0.693, 0.785]} & 0.659 {\scriptsize [0.598, 0.717]} & 0.847 {\scriptsize [0.795, 0.896]} \\
Qwen-3-14B & 0.910 {\scriptsize [0.884, 0.934]} & 0.802 {\scriptsize [0.766, 0.836]} & 0.769 {\scriptsize [0.722, 0.812]} & 0.708 {\scriptsize [0.649, 0.765]} & 0.842 {\scriptsize [0.789, 0.891]} \\
\midrule
\multicolumn{6}{l}{\textit{Linear probe, 5-fold cross-validation over all 2,000 sentences}} \\
Gemma-4-31B-it, last & 0.971 {\scriptsize [0.964, 0.977]} & 0.913 {\scriptsize [0.900, 0.925]} & 0.890 {\scriptsize [0.874, 0.906]} & 0.896 {\scriptsize [0.873, 0.916]} & 0.886 {\scriptsize [0.863, 0.907]} \\
Gemma-4-31B-it, mean\rlap{$^{\dagger}$} & 0.973 {\scriptsize [0.967, 0.979]} & 0.915 {\scriptsize [0.903, 0.927]} & 0.894 {\scriptsize [0.878, 0.910]} & 0.895 {\scriptsize [0.874, 0.916]} & 0.893 {\scriptsize [0.871, 0.914]} \\
Gemma-4-31B-it, max & 0.977 {\scriptsize [0.971, 0.982]} & 0.920 {\scriptsize [0.908, 0.932]} & 0.900 {\scriptsize [0.884, 0.915]} & 0.905 {\scriptsize [0.884, 0.924]} & 0.895 {\scriptsize [0.874, 0.916]} \\
\addlinespace[2pt]
Llama-3.1-8B, last & 0.964 {\scriptsize [0.956, 0.971]} & 0.904 {\scriptsize [0.891, 0.917]} & 0.880 {\scriptsize [0.862, 0.896]} & 0.887 {\scriptsize [0.864, 0.908]} & 0.873 {\scriptsize [0.849, 0.896]} \\
Llama-3.1-8B, mean & 0.968 {\scriptsize [0.961, 0.975]} & 0.912 {\scriptsize [0.900, 0.925]} & 0.890 {\scriptsize [0.874, 0.906]} & 0.891 {\scriptsize [0.869, 0.911]} & 0.891 {\scriptsize [0.868, 0.912]} \\
Llama-3.1-8B, max\rlap{$^{\dagger}$} & 0.972 {\scriptsize [0.966, 0.979]} & 0.921 {\scriptsize [0.909, 0.933]} & 0.902 {\scriptsize [0.887, 0.917]} & 0.897 {\scriptsize [0.876, 0.917]} & 0.908 {\scriptsize [0.887, 0.928]} \\
\addlinespace[2pt]
Qwen-3-8B, last & 0.963 {\scriptsize [0.955, 0.970]} & 0.903 {\scriptsize [0.890, 0.916]} & 0.879 {\scriptsize [0.862, 0.896]} & 0.880 {\scriptsize [0.857, 0.902]} & 0.878 {\scriptsize [0.854, 0.900]} \\
Qwen-3-8B, mean & 0.970 {\scriptsize [0.963, 0.976]} & 0.911 {\scriptsize [0.898, 0.924]} & 0.889 {\scriptsize [0.872, 0.905]} & 0.892 {\scriptsize [0.870, 0.913]} & 0.886 {\scriptsize [0.863, 0.907]} \\
Qwen-3-8B, max\rlap{$^{\dagger}$} & 0.976 {\scriptsize [0.970, 0.981]} & 0.922 {\scriptsize [0.910, 0.934]} & 0.902 {\scriptsize [0.887, 0.917]} & 0.907 {\scriptsize [0.886, 0.927]} & 0.898 {\scriptsize [0.877, 0.918]} \\
\addlinespace[2pt]
Qwen-3-14B, last & 0.966 {\scriptsize [0.959, 0.973]} & 0.905 {\scriptsize [0.892, 0.917]} & 0.880 {\scriptsize [0.863, 0.896]} & 0.887 {\scriptsize [0.864, 0.908]} & 0.874 {\scriptsize [0.852, 0.897]} \\
Qwen-3-14B, mean & 0.969 {\scriptsize [0.962, 0.975]} & 0.917 {\scriptsize [0.905, 0.929]} & 0.896 {\scriptsize [0.880, 0.912]} & 0.899 {\scriptsize [0.877, 0.919]} & 0.894 {\scriptsize [0.873, 0.915]} \\
Qwen-3-14B, max\rlap{$^{\dagger}$} & 0.976 {\scriptsize [0.970, 0.982]} & 0.920 {\scriptsize [0.908, 0.932]} & 0.900 {\scriptsize [0.884, 0.915]} & 0.901 {\scriptsize [0.881, 0.922]} & 0.898 {\scriptsize [0.877, 0.918]} \\
\midrule
\multicolumn{6}{l}{\textit{Linear probe, RFM split (train 1,000, test $n = 500$)}} \\
Gemma-4-31B-it, last & 0.957 {\scriptsize [0.940, 0.972]} & 0.904 {\scriptsize [0.878, 0.928]} & 0.878 {\scriptsize [0.842, 0.910]} & 0.878 {\scriptsize [0.831, 0.922]} & 0.878 {\scriptsize [0.829, 0.922]} \\
Gemma-4-31B-it, mean\rlap{$^{\dagger}$} & 0.955 {\scriptsize [0.937, 0.971]} & 0.890 {\scriptsize [0.862, 0.916]} & 0.864 {\scriptsize [0.826, 0.898]} & 0.841 {\scriptsize [0.789, 0.889]} & 0.888 {\scriptsize [0.841, 0.931]} \\
Gemma-4-31B-it, max & 0.965 {\scriptsize [0.951, 0.978]} & 0.890 {\scriptsize [0.862, 0.916]} & 0.861 {\scriptsize [0.822, 0.896]} & 0.854 {\scriptsize [0.803, 0.901]} & 0.867 {\scriptsize [0.819, 0.913]} \\
\addlinespace[2pt]
Llama-3.1-8B, last & 0.950 {\scriptsize [0.931, 0.966]} & 0.890 {\scriptsize [0.862, 0.916]} & 0.859 {\scriptsize [0.820, 0.894]} & 0.865 {\scriptsize [0.816, 0.912]} & 0.852 {\scriptsize [0.802, 0.900]} \\
Llama-3.1-8B, mean & 0.948 {\scriptsize [0.928, 0.966]} & 0.884 {\scriptsize [0.856, 0.912]} & 0.854 {\scriptsize [0.815, 0.889]} & 0.845 {\scriptsize [0.794, 0.894]} & 0.862 {\scriptsize [0.812, 0.909]} \\
Llama-3.1-8B, max\rlap{$^{\dagger}$} & 0.959 {\scriptsize [0.943, 0.974]} & 0.896 {\scriptsize [0.868, 0.922]} & 0.869 {\scriptsize [0.831, 0.902]} & 0.860 {\scriptsize [0.810, 0.907]} & 0.878 {\scriptsize [0.830, 0.921]} \\
\addlinespace[2pt]
Qwen-3-8B, last & 0.938 {\scriptsize [0.916, 0.959]} & 0.876 {\scriptsize [0.846, 0.904]} & 0.840 {\scriptsize [0.799, 0.878]} & 0.849 {\scriptsize [0.797, 0.897]} & 0.832 {\scriptsize [0.779, 0.882]} \\
Qwen-3-8B, mean & 0.950 {\scriptsize [0.931, 0.967]} & 0.884 {\scriptsize [0.854, 0.912]} & 0.856 {\scriptsize [0.818, 0.892]} & 0.832 {\scriptsize [0.780, 0.880]} & 0.883 {\scriptsize [0.835, 0.926]} \\
Qwen-3-8B, max\rlap{$^{\dagger}$} & 0.962 {\scriptsize [0.946, 0.976]} & 0.910 {\scriptsize [0.884, 0.934]} & 0.886 {\scriptsize [0.851, 0.917]} & 0.879 {\scriptsize [0.833, 0.922]} & 0.893 {\scriptsize [0.848, 0.934]} \\
\addlinespace[2pt]
Qwen-3-14B, last & 0.949 {\scriptsize [0.927, 0.967]} & 0.892 {\scriptsize [0.864, 0.918]} & 0.865 {\scriptsize [0.827, 0.899]} & 0.848 {\scriptsize [0.797, 0.896]} & 0.883 {\scriptsize [0.836, 0.927]} \\
Qwen-3-14B, mean & 0.951 {\scriptsize [0.931, 0.968]} & 0.892 {\scriptsize [0.864, 0.918]} & 0.865 {\scriptsize [0.827, 0.899]} & 0.848 {\scriptsize [0.796, 0.895]} & 0.883 {\scriptsize [0.836, 0.926]} \\
Qwen-3-14B, max\rlap{$^{\dagger}$} & 0.968 {\scriptsize [0.953, 0.980]} & 0.912 {\scriptsize [0.886, 0.936]} & 0.888 {\scriptsize [0.852, 0.919]} & 0.888 {\scriptsize [0.842, 0.930]} & 0.888 {\scriptsize [0.841, 0.930]} \\
\midrule
\multicolumn{6}{l}{\textit{Surface and embedding baselines, 5-fold CV over 2,000 (mean $\pm$ sd over folds)}} \\
Dictionary + LR & 0.734 {\scriptsize $\pm$0.011} & 0.670 {\scriptsize $\pm$0.011} & 0.356 {\scriptsize $\pm$0.023} & 0.821 {\scriptsize $\pm$0.042} & 0.228 {\scriptsize $\pm$0.017} \\
LDA + LR & 0.805 {\scriptsize $\pm$0.017} & 0.731 {\scriptsize $\pm$0.012} & 0.616 {\scriptsize $\pm$0.029} & 0.723 {\scriptsize $\pm$0.035} & 0.540 {\scriptsize $\pm$0.051} \\
TF-IDF + LR & 0.943 {\scriptsize $\pm$0.008} & 0.865 {\scriptsize $\pm$0.024} & 0.823 {\scriptsize $\pm$0.033} & 0.863 {\scriptsize $\pm$0.025} & 0.787 {\scriptsize $\pm$0.042} \\
Embedding + LR & 0.977 {\scriptsize $\pm$0.004} & 0.925 {\scriptsize $\pm$0.019} & 0.909 {\scriptsize $\pm$0.021} & 0.882 {\scriptsize $\pm$0.032} & 0.939 {\scriptsize $\pm$0.021} \\
\midrule
\multicolumn{6}{l}{\textit{Fine-tuned classifiers (Schimanski et al., 2024, mean $\pm$ sd as published)}} \\
SocRoBERTa & & 0.934 {\scriptsize $\pm$0.014} & 0.919 {\scriptsize $\pm$0.018} & 0.904 {\scriptsize $\pm$0.035} & 0.937 {\scriptsize $\pm$0.029} \\
SocDistilRoBERTa & & 0.932 {\scriptsize $\pm$0.015} & 0.912 {\scriptsize $\pm$0.019} & 0.913 {\scriptsize $\pm$0.027} & 0.913 {\scriptsize $\pm$0.025} \\
\bottomrule
\end{tabular}

  \caption{All methods on the Social pillar. Protocols and notation as in Table \ref{tab:full_environmental}.}
  \label{tab:full_social}
\end{table*}

\begin{table*}[tp]
  \centering
  \footnotesize
  \begin{tabular}{@{}l@{\hspace{6pt}}c@{\hspace{6pt}}c@{\hspace{6pt}}c@{\hspace{6pt}}c@{\hspace{6pt}}c@{}}
\toprule
Method & AUC & Acc & F1 & Precision & Recall \\
\midrule
\multicolumn{6}{l}{\textit{Model's own answer (test split, $n = 500$)}} \\
Gemma-4-31B-it & 0.834 {\scriptsize [0.796, 0.870]} & 0.756 {\scriptsize [0.718, 0.792]} & 0.623 {\scriptsize [0.560, 0.683]} & 0.505 {\scriptsize [0.437, 0.574]} & 0.815 {\scriptsize [0.744, 0.881]} \\
Llama-3.1-8B & 0.842 {\scriptsize [0.801, 0.881]} & 0.820 {\scriptsize [0.786, 0.852]} & 0.583 {\scriptsize [0.502, 0.658]} & 0.685 {\scriptsize [0.589, 0.782]} & 0.508 {\scriptsize [0.420, 0.597]} \\
Qwen-3-8B & 0.830 {\scriptsize [0.782, 0.875]} & 0.842 {\scriptsize [0.810, 0.874]} & 0.664 {\scriptsize [0.592, 0.729]} & 0.703 {\scriptsize [0.618, 0.785]} & 0.629 {\scriptsize [0.544, 0.713]} \\
Qwen-3-14B & 0.861 {\scriptsize [0.821, 0.896]} & 0.800 {\scriptsize [0.764, 0.834]} & 0.648 {\scriptsize [0.579, 0.709]} & 0.575 {\scriptsize [0.497, 0.650]} & 0.742 {\scriptsize [0.664, 0.816]} \\
\midrule
\multicolumn{6}{l}{\textit{RFM, first concept vector, frozen threshold (test split, $n = 500$)}} \\
Gemma-4-31B-it & 0.811 {\scriptsize [0.766, 0.853]} & 0.770 {\scriptsize [0.732, 0.808]} & 0.531 {\scriptsize [0.451, 0.604]} & 0.537 {\scriptsize [0.448, 0.627]} & 0.524 {\scriptsize [0.437, 0.612]} \\
Llama-3.1-8B & 0.877 {\scriptsize [0.839, 0.911]} & 0.842 {\scriptsize [0.810, 0.872]} & 0.622 {\scriptsize [0.543, 0.697]} & 0.765 {\scriptsize [0.673, 0.853]} & 0.524 {\scriptsize [0.438, 0.611]} \\
Qwen-3-8B & 0.863 {\scriptsize [0.823, 0.900]} & 0.824 {\scriptsize [0.790, 0.856]} & 0.654 {\scriptsize [0.581, 0.717]} & 0.638 {\scriptsize [0.553, 0.720]} & 0.669 {\scriptsize [0.584, 0.748]} \\
Qwen-3-14B & 0.872 {\scriptsize [0.832, 0.910]} & 0.846 {\scriptsize [0.814, 0.876]} & 0.621 {\scriptsize [0.538, 0.696]} & 0.797 {\scriptsize [0.705, 0.884]} & 0.508 {\scriptsize [0.421, 0.597]} \\
\midrule
\multicolumn{6}{l}{\textit{RFM, mean of 5 concept vectors, frozen threshold (test split, $n = 500$)}} \\
Gemma-4-31B-it & 0.855 {\scriptsize [0.815, 0.893]} & 0.810 {\scriptsize [0.774, 0.844]} & 0.649 {\scriptsize [0.581, 0.713]} & 0.599 {\scriptsize [0.519, 0.677]} & 0.710 {\scriptsize [0.628, 0.789]} \\
Llama-3.1-8B & 0.881 {\scriptsize [0.847, 0.913]} & 0.840 {\scriptsize [0.808, 0.872]} & 0.675 {\scriptsize [0.604, 0.738]} & 0.680 {\scriptsize [0.595, 0.759]} & 0.669 {\scriptsize [0.587, 0.752]} \\
Qwen-3-8B & 0.847 {\scriptsize [0.806, 0.885]} & 0.802 {\scriptsize [0.766, 0.836]} & 0.407 {\scriptsize [0.308, 0.497]} & 0.791 {\scriptsize [0.659, 0.906]} & 0.274 {\scriptsize [0.196, 0.353]} \\
Qwen-3-14B & 0.871 {\scriptsize [0.835, 0.903]} & 0.810 {\scriptsize [0.776, 0.844]} & 0.541 {\scriptsize [0.456, 0.621]} & 0.675 {\scriptsize [0.571, 0.775]} & 0.452 {\scriptsize [0.364, 0.541]} \\
\midrule
\multicolumn{6}{l}{\textit{Linear probe, 5-fold cross-validation over all 2,000 sentences}} \\
Gemma-4-31B-it, last & 0.915 {\scriptsize [0.901, 0.927]} & 0.852 {\scriptsize [0.837, 0.868]} & 0.707 {\scriptsize [0.674, 0.738]} & 0.755 {\scriptsize [0.717, 0.793]} & 0.665 {\scriptsize [0.626, 0.705]} \\
Gemma-4-31B-it, mean & 0.896 {\scriptsize [0.879, 0.910]} & 0.860 {\scriptsize [0.845, 0.875]} & 0.723 {\scriptsize [0.692, 0.753]} & 0.773 {\scriptsize [0.734, 0.808]} & 0.680 {\scriptsize [0.642, 0.720]} \\
Gemma-4-31B-it, max\rlap{$^{\dagger}$} & 0.911 {\scriptsize [0.896, 0.925]} & 0.861 {\scriptsize [0.845, 0.876]} & 0.729 {\scriptsize [0.697, 0.759]} & 0.765 {\scriptsize [0.726, 0.801]} & 0.697 {\scriptsize [0.657, 0.736]} \\
\addlinespace[2pt]
Llama-3.1-8B, last\rlap{$^{\dagger}$} & 0.913 {\scriptsize [0.898, 0.926]} & 0.860 {\scriptsize [0.844, 0.875]} & 0.718 {\scriptsize [0.685, 0.750]} & 0.781 {\scriptsize [0.741, 0.818]} & 0.665 {\scriptsize [0.625, 0.705]} \\
Llama-3.1-8B, mean & 0.898 {\scriptsize [0.881, 0.912]} & 0.857 {\scriptsize [0.841, 0.872]} & 0.707 {\scriptsize [0.674, 0.738]} & 0.786 {\scriptsize [0.746, 0.822]} & 0.643 {\scriptsize [0.604, 0.684]} \\
Llama-3.1-8B, max & 0.908 {\scriptsize [0.892, 0.922]} & 0.864 {\scriptsize [0.849, 0.879]} & 0.733 {\scriptsize [0.703, 0.763]} & 0.775 {\scriptsize [0.737, 0.812]} & 0.697 {\scriptsize [0.659, 0.736]} \\
\addlinespace[2pt]
Qwen-3-8B, last & 0.907 {\scriptsize [0.891, 0.920]} & 0.860 {\scriptsize [0.845, 0.875]} & 0.721 {\scriptsize [0.689, 0.752]} & 0.777 {\scriptsize [0.739, 0.814]} & 0.673 {\scriptsize [0.632, 0.712]} \\
Qwen-3-8B, mean & 0.895 {\scriptsize [0.876, 0.909]} & 0.846 {\scriptsize [0.830, 0.862]} & 0.685 {\scriptsize [0.652, 0.718]} & 0.763 {\scriptsize [0.719, 0.800]} & 0.624 {\scriptsize [0.583, 0.665]} \\
Qwen-3-8B, max\rlap{$^{\dagger}$} & 0.909 {\scriptsize [0.893, 0.922]} & 0.859 {\scriptsize [0.843, 0.874]} & 0.726 {\scriptsize [0.694, 0.756]} & 0.758 {\scriptsize [0.719, 0.795]} & 0.697 {\scriptsize [0.657, 0.736]} \\
\addlinespace[2pt]
Qwen-3-14B, last & 0.907 {\scriptsize [0.892, 0.921]} & 0.858 {\scriptsize [0.843, 0.873]} & 0.714 {\scriptsize [0.683, 0.746]} & 0.774 {\scriptsize [0.736, 0.812]} & 0.664 {\scriptsize [0.625, 0.703]} \\
Qwen-3-14B, mean & 0.900 {\scriptsize [0.883, 0.914]} & 0.860 {\scriptsize [0.845, 0.875]} & 0.715 {\scriptsize [0.683, 0.746]} & 0.786 {\scriptsize [0.748, 0.823]} & 0.656 {\scriptsize [0.616, 0.696]} \\
Qwen-3-14B, max\rlap{$^{\dagger}$} & 0.910 {\scriptsize [0.895, 0.924]} & 0.863 {\scriptsize [0.848, 0.878]} & 0.730 {\scriptsize [0.699, 0.760]} & 0.775 {\scriptsize [0.738, 0.813]} & 0.689 {\scriptsize [0.651, 0.729]} \\
\midrule
\multicolumn{6}{l}{\textit{Linear probe, RFM split (train 1,000, test $n = 500$)}} \\
Gemma-4-31B-it, last & 0.904 {\scriptsize [0.871, 0.933]} & 0.858 {\scriptsize [0.826, 0.888]} & 0.710 {\scriptsize [0.643, 0.770]} & 0.719 {\scriptsize [0.636, 0.796]} & 0.702 {\scriptsize [0.621, 0.781]} \\
Gemma-4-31B-it, mean & 0.900 {\scriptsize [0.867, 0.931]} & 0.870 {\scriptsize [0.840, 0.898]} & 0.721 {\scriptsize [0.651, 0.782]} & 0.771 {\scriptsize [0.690, 0.848]} & 0.677 {\scriptsize [0.595, 0.757]} \\
Gemma-4-31B-it, max\rlap{$^{\dagger}$} & 0.914 {\scriptsize [0.885, 0.941]} & 0.860 {\scriptsize [0.830, 0.890]} & 0.715 {\scriptsize [0.648, 0.776]} & 0.721 {\scriptsize [0.642, 0.800]} & 0.710 {\scriptsize [0.630, 0.788]} \\
\addlinespace[2pt]
Llama-3.1-8B, last\rlap{$^{\dagger}$} & 0.907 {\scriptsize [0.874, 0.937]} & 0.876 {\scriptsize [0.846, 0.904]} & 0.730 {\scriptsize [0.663, 0.791]} & 0.792 {\scriptsize [0.713, 0.867]} & 0.677 {\scriptsize [0.595, 0.757]} \\
Llama-3.1-8B, mean & 0.890 {\scriptsize [0.854, 0.923]} & 0.870 {\scriptsize [0.840, 0.898]} & 0.723 {\scriptsize [0.654, 0.784]} & 0.766 {\scriptsize [0.686, 0.839]} & 0.685 {\scriptsize [0.602, 0.766]} \\
Llama-3.1-8B, max & 0.905 {\scriptsize [0.870, 0.936]} & 0.874 {\scriptsize [0.844, 0.902]} & 0.741 {\scriptsize [0.675, 0.799]} & 0.756 {\scriptsize [0.678, 0.829]} & 0.726 {\scriptsize [0.646, 0.803]} \\
\addlinespace[2pt]
Qwen-3-8B, last & 0.889 {\scriptsize [0.853, 0.920]} & 0.840 {\scriptsize [0.808, 0.870]} & 0.669 {\scriptsize [0.596, 0.735]} & 0.686 {\scriptsize [0.602, 0.769]} & 0.653 {\scriptsize [0.568, 0.736]} \\
Qwen-3-8B, mean & 0.890 {\scriptsize [0.854, 0.923]} & 0.854 {\scriptsize [0.822, 0.886]} & 0.697 {\scriptsize [0.626, 0.761]} & 0.718 {\scriptsize [0.634, 0.797]} & 0.677 {\scriptsize [0.593, 0.760]} \\
Qwen-3-8B, max\rlap{$^{\dagger}$} & 0.902 {\scriptsize [0.868, 0.932]} & 0.866 {\scriptsize [0.836, 0.894]} & 0.724 {\scriptsize [0.657, 0.784]} & 0.739 {\scriptsize [0.659, 0.817]} & 0.710 {\scriptsize [0.627, 0.786]} \\
\addlinespace[2pt]
Qwen-3-14B, last & 0.894 {\scriptsize [0.858, 0.928]} & 0.864 {\scriptsize [0.832, 0.894]} & 0.717 {\scriptsize [0.649, 0.778]} & 0.741 {\scriptsize [0.661, 0.820]} & 0.694 {\scriptsize [0.612, 0.773]} \\
Qwen-3-14B, mean & 0.889 {\scriptsize [0.854, 0.920]} & 0.864 {\scriptsize [0.834, 0.894]} & 0.717 {\scriptsize [0.648, 0.779]} & 0.741 {\scriptsize [0.660, 0.820]} & 0.694 {\scriptsize [0.611, 0.774]} \\
Qwen-3-14B, max\rlap{$^{\dagger}$} & 0.897 {\scriptsize [0.862, 0.929]} & 0.868 {\scriptsize [0.838, 0.898]} & 0.725 {\scriptsize [0.658, 0.786]} & 0.750 {\scriptsize [0.670, 0.827]} & 0.702 {\scriptsize [0.620, 0.780]} \\
\midrule
\multicolumn{6}{l}{\textit{Surface and embedding baselines, 5-fold CV over 2,000 (mean $\pm$ sd over folds)}} \\
Dictionary + LR & 0.768 {\scriptsize $\pm$0.023} & 0.741 {\scriptsize $\pm$0.007} & 0.128 {\scriptsize $\pm$0.026} & 0.673 {\scriptsize $\pm$0.116} & 0.071 {\scriptsize $\pm$0.015} \\
LDA + LR & 0.780 {\scriptsize $\pm$0.024} & 0.769 {\scriptsize $\pm$0.011} & 0.373 {\scriptsize $\pm$0.096} & 0.683 {\scriptsize $\pm$0.019} & 0.266 {\scriptsize $\pm$0.085} \\
TF-IDF + LR & 0.894 {\scriptsize $\pm$0.018} & 0.856 {\scriptsize $\pm$0.008} & 0.695 {\scriptsize $\pm$0.024} & 0.807 {\scriptsize $\pm$0.021} & 0.612 {\scriptsize $\pm$0.038} \\
Embedding + LR & 0.918 {\scriptsize $\pm$0.012} & 0.836 {\scriptsize $\pm$0.007} & 0.731 {\scriptsize $\pm$0.011} & 0.653 {\scriptsize $\pm$0.022} & 0.833 {\scriptsize $\pm$0.040} \\
\midrule
\multicolumn{6}{l}{\textit{Fine-tuned classifiers (Schimanski et al., 2024, mean $\pm$ sd as published)}} \\
GovRoBERTa & & 0.896 {\scriptsize $\pm$0.011} & 0.785 {\scriptsize $\pm$0.026} & 0.856 {\scriptsize $\pm$0.018} & 0.725 {\scriptsize $\pm$0.038} \\
GovDistilRoBERTa & & 0.897 {\scriptsize $\pm$0.011} & 0.789 {\scriptsize $\pm$0.016} & 0.854 {\scriptsize $\pm$0.047} & 0.734 {\scriptsize $\pm$0.015} \\
\bottomrule
\end{tabular}

  \caption{All methods on the Governance pillar. Protocols and notation as in Table \ref{tab:full_environmental}.}
  \label{tab:full_governance}
\end{table*}

\section{RFM threshold: swept against frozen}
\label{app:threshold}
Table \ref{tab:rfm_swept} compares the RFM threshold swept on the test split with the threshold fitted on validation and frozen.

\begin{table*}[tp]
  \centering
  \footnotesize
  \begin{tabular}{@{}llrrrrrr@{}}
\toprule
 & & \multicolumn{3}{c}{Accuracy} & \multicolumn{3}{c}{F1} \\
\cmidrule(lr){3-5} \cmidrule(lr){6-8}
Model & Vector & Swept & Frozen & Gap & Swept & Frozen & Gap \\
\midrule
\multicolumn{8}{l}{\textit{Environmental}} \\
Gemma-4-31B-it & first & 0.938 & 0.936 & 0.002 & 0.916 & 0.912 & 0.004 \\
 & mean of 5 & 0.942 & 0.940 & 0.002 & 0.919 & 0.915 & 0.004 \\
Llama-3.1-8B & first & 0.898 & 0.894 & 0.004 & 0.860 & 0.845 & 0.015 \\
 & mean of 5 & 0.886 & 0.872 & 0.014 & 0.850 & 0.814 & 0.036 \\
Qwen-3-8B & first & 0.898 & 0.892 & 0.006 & 0.848 & 0.850 & $-$0.002 \\
 & mean of 5 & 0.846 & 0.844 & 0.002 & 0.757 & 0.755 & 0.002 \\
Qwen-3-14B & first & 0.942 & 0.936 & 0.006 & 0.916 & 0.910 & 0.006 \\
 & mean of 5 & 0.940 & 0.938 & 0.002 & 0.916 & 0.913 & 0.003 \\
\midrule
\multicolumn{8}{l}{\textit{Social}} \\
Gemma-4-31B-it & first & 0.884 & 0.868 & 0.016 & 0.863 & 0.832 & 0.031 \\
 & mean of 5 & 0.836 & 0.834 & 0.002 & 0.777 & 0.773 & 0.004 \\
Llama-3.1-8B & first & 0.860 & 0.852 & 0.008 & 0.821 & 0.812 & 0.009 \\
 & mean of 5 & 0.828 & 0.824 & 0.004 & 0.797 & 0.774 & 0.023 \\
Qwen-3-8B & first & 0.836 & 0.818 & 0.018 & 0.780 & 0.787 & $-$0.007 \\
 & mean of 5 & 0.774 & 0.768 & 0.006 & 0.734 & 0.741 & $-$0.007 \\
Qwen-3-14B & first & 0.858 & 0.854 & 0.004 & 0.820 & 0.807 & 0.013 \\
 & mean of 5 & 0.832 & 0.802 & 0.030 & 0.787 & 0.769 & 0.018 \\
\midrule
\multicolumn{8}{l}{\textit{Governance}} \\
Gemma-4-31B-it & first & 0.808 & 0.770 & 0.038 & 0.422 & 0.531 & $-$0.109 \\
 & mean of 5 & 0.842 & 0.810 & 0.032 & 0.643 & 0.649 & $-$0.006 \\
Llama-3.1-8B & first & 0.854 & 0.842 & 0.012 & 0.707 & 0.622 & 0.085 \\
 & mean of 5 & 0.860 & 0.840 & 0.020 & 0.667 & 0.675 & $-$0.008 \\
Qwen-3-8B & first & 0.842 & 0.824 & 0.018 & 0.607 & 0.654 & $-$0.047 \\
 & mean of 5 & 0.828 & 0.802 & 0.026 & 0.659 & 0.407 & 0.252 \\
Qwen-3-14B & first & 0.858 & 0.846 & 0.012 & 0.679 & 0.621 & 0.058 \\
 & mean of 5 & 0.830 & 0.810 & 0.020 & 0.667 & 0.541 & 0.126 \\
\bottomrule
\end{tabular}

  \caption{RFM accuracy and F1 with the decision threshold swept on the test split against the threshold fitted on validation and frozen. Gap: swept minus frozen, the optimism of choosing the threshold on the test labels. Test split, $n = 500$.}
  \label{tab:rfm_swept}
\end{table*}

\section{Paired tests}
\label{app:paired}
Table \ref{tab:paired} reports the paired tests of Section \ref{sec:stats} for every model-pillar cell.

\begin{table*}[tp]
  \centering
  \footnotesize
  \begin{tabular}{@{}llccrrr@{}}
\toprule
 & & & & \multicolumn{3}{c}{Holm-adjusted $p$} \\
\cmidrule(lr){5-7}
Probe & Versus & $\Delta$Acc [95\% CI] & $\Delta$AUC [95\% CI] & Acc & AUC & McNemar \\
\midrule
\multicolumn{7}{l}{\textit{Environmental}} \\
Gemma-4-31B-it (last) & own answer & 0.012 {\scriptsize [$-$0.006, 0.030]} & \phantom{$-$}0.016 {\scriptsize \phantom{$-$}[0.007, 0.025]} & 0.357 & 0.001 & 0.413 \\
 & RFM first & 0.006 {\scriptsize [$-$0.010, 0.022]} & \phantom{$-$}0.014 {\scriptsize \phantom{$-$}[0.005, 0.024]} & 1.000 & 0.002 & 1.000 \\
 & RFM mean of 5 & 0.002 {\scriptsize [$-$0.012, 0.016]} & \phantom{$-$}0.005 {\scriptsize \phantom{$-$}[0.000, 0.010]} & 1.000 & 0.107 & 1.000 \\
\addlinespace[2pt]
Llama-3.1-8B (last) & own answer & 0.074 {\scriptsize \phantom{$-$}[0.042, 0.106]} & \phantom{$-$}0.014 {\scriptsize \phantom{$-$}[0.002, 0.026]} & 0.001 & 0.071 & $<$0.001 \\
 & RFM first & 0.044 {\scriptsize \phantom{$-$}[0.018, 0.072]} & \phantom{$-$}0.021 {\scriptsize \phantom{$-$}[0.008, 0.034]} & 0.011 & 0.002 & 0.013 \\
 & RFM mean of 5 & 0.066 {\scriptsize \phantom{$-$}[0.036, 0.098]} & \phantom{$-$}0.040 {\scriptsize \phantom{$-$}[0.022, 0.060]} & 0.001 & 0.001 & $<$0.001 \\
\addlinespace[2pt]
Qwen-3-8B (last) & own answer & 0.032 {\scriptsize \phantom{$-$}[0.010, 0.054]} & \phantom{$-$}0.005 {\scriptsize [$-$0.003, 0.014]} & 0.019 & 0.455 & 0.028 \\
 & RFM first & 0.060 {\scriptsize \phantom{$-$}[0.034, 0.088]} & \phantom{$-$}0.019 {\scriptsize \phantom{$-$}[0.009, 0.031]} & 0.001 & 0.001 & $<$0.001 \\
 & RFM mean of 5 & 0.108 {\scriptsize \phantom{$-$}[0.076, 0.142]} & \phantom{$-$}0.073 {\scriptsize \phantom{$-$}[0.051, 0.097]} & 0.001 & 0.001 & $<$0.001 \\
\addlinespace[2pt]
Qwen-3-14B (last) & own answer & 0.026 {\scriptsize \phantom{$-$}[0.004, 0.048]} & \phantom{$-$}0.003 {\scriptsize [$-$0.007, 0.013]} & 0.089 & 0.544 & 0.105 \\
 & RFM first & 0.006 {\scriptsize [$-$0.012, 0.024]} & $-$0.002 {\scriptsize [$-$0.007, 0.001]} & 1.000 & 0.251 & 1.000 \\
 & RFM mean of 5 & 0.004 {\scriptsize [$-$0.016, 0.024]} & \phantom{$-$}0.001 {\scriptsize [$-$0.007, 0.008]} & 1.000 & 0.789 & 1.000 \\
\midrule
\multicolumn{7}{l}{\textit{Social}} \\
Gemma-4-31B-it (mean) & own answer & 0.086 {\scriptsize \phantom{$-$}[0.052, 0.120]} & \phantom{$-$}0.095 {\scriptsize \phantom{$-$}[0.061, 0.132]} & 0.001 & 0.001 & $<$0.001 \\
 & RFM first & 0.022 {\scriptsize [$-$0.006, 0.050]} & \phantom{$-$}0.012 {\scriptsize [$-$0.003, 0.028]} & 0.522 & 0.251 & 0.610 \\
 & RFM mean of 5 & 0.056 {\scriptsize \phantom{$-$}[0.026, 0.088]} & \phantom{$-$}0.033 {\scriptsize \phantom{$-$}[0.014, 0.053]} & 0.001 & 0.007 & 0.005 \\
\addlinespace[2pt]
Llama-3.1-8B (max) & own answer & 0.074 {\scriptsize \phantom{$-$}[0.040, 0.108]} & \phantom{$-$}0.067 {\scriptsize \phantom{$-$}[0.042, 0.094]} & 0.001 & 0.001 & $<$0.001 \\
 & RFM first & 0.044 {\scriptsize \phantom{$-$}[0.012, 0.076]} & \phantom{$-$}0.038 {\scriptsize \phantom{$-$}[0.018, 0.058]} & 0.057 & 0.002 & 0.064 \\
 & RFM mean of 5 & 0.072 {\scriptsize \phantom{$-$}[0.038, 0.106]} & \phantom{$-$}0.057 {\scriptsize \phantom{$-$}[0.035, 0.081]} & 0.001 & 0.001 & $<$0.001 \\
\addlinespace[2pt]
Qwen-3-8B (max) & own answer & 0.074 {\scriptsize \phantom{$-$}[0.042, 0.106]} & \phantom{$-$}0.025 {\scriptsize \phantom{$-$}[0.008, 0.043]} & 0.001 & 0.019 & $<$0.001 \\
 & RFM first & 0.092 {\scriptsize \phantom{$-$}[0.058, 0.126]} & \phantom{$-$}0.055 {\scriptsize \phantom{$-$}[0.034, 0.077]} & 0.001 & 0.001 & $<$0.001 \\
 & RFM mean of 5 & 0.142 {\scriptsize \phantom{$-$}[0.102, 0.182]} & \phantom{$-$}0.121 {\scriptsize \phantom{$-$}[0.090, 0.155]} & 0.001 & 0.001 & $<$0.001 \\
\addlinespace[2pt]
Qwen-3-14B (max) & own answer & 0.102 {\scriptsize \phantom{$-$}[0.068, 0.136]} & \phantom{$-$}0.075 {\scriptsize \phantom{$-$}[0.051, 0.101]} & 0.001 & 0.001 & $<$0.001 \\
 & RFM first & 0.058 {\scriptsize \phantom{$-$}[0.030, 0.086]} & \phantom{$-$}0.033 {\scriptsize \phantom{$-$}[0.018, 0.049]} & 0.002 & 0.001 & $<$0.001 \\
 & RFM mean of 5 & 0.110 {\scriptsize \phantom{$-$}[0.074, 0.146]} & \phantom{$-$}0.058 {\scriptsize \phantom{$-$}[0.037, 0.081]} & 0.001 & 0.001 & $<$0.001 \\
\midrule
\multicolumn{7}{l}{\textit{Governance}} \\
Gemma-4-31B-it (max) & own answer & 0.104 {\scriptsize \phantom{$-$}[0.064, 0.144]} & \phantom{$-$}0.080 {\scriptsize \phantom{$-$}[0.047, 0.114]} & 0.001 & 0.001 & $<$0.001 \\
 & RFM first & 0.090 {\scriptsize \phantom{$-$}[0.050, 0.130]} & \phantom{$-$}0.103 {\scriptsize \phantom{$-$}[0.063, 0.144]} & 0.001 & 0.001 & $<$0.001 \\
 & RFM mean of 5 & 0.050 {\scriptsize \phantom{$-$}[0.014, 0.084]} & \phantom{$-$}0.058 {\scriptsize \phantom{$-$}[0.028, 0.089]} & 0.027 & 0.001 & 0.029 \\
\addlinespace[2pt]
Llama-3.1-8B (last) & own answer & 0.056 {\scriptsize \phantom{$-$}[0.022, 0.090]} & \phantom{$-$}0.064 {\scriptsize \phantom{$-$}[0.029, 0.101]} & 0.003 & 0.003 & 0.009 \\
 & RFM first & 0.034 {\scriptsize \phantom{$-$}[0.004, 0.066]} & \phantom{$-$}0.030 {\scriptsize \phantom{$-$}[0.001, 0.060]} & 0.170 & 0.158 & 0.215 \\
 & RFM mean of 5 & 0.036 {\scriptsize \phantom{$-$}[0.002, 0.070]} & \phantom{$-$}0.025 {\scriptsize [$-$0.006, 0.058]} & 0.140 & 0.349 & 0.161 \\
\addlinespace[2pt]
Qwen-3-8B (max) & own answer & 0.024 {\scriptsize [$-$0.010, 0.058]} & \phantom{$-$}0.072 {\scriptsize \phantom{$-$}[0.032, 0.113]} & 0.357 & 0.003 & 0.413 \\
 & RFM first & 0.042 {\scriptsize \phantom{$-$}[0.008, 0.074]} & \phantom{$-$}0.038 {\scriptsize \phantom{$-$}[0.009, 0.067]} & 0.110 & 0.049 & 0.112 \\
 & RFM mean of 5 & 0.064 {\scriptsize \phantom{$-$}[0.026, 0.102]} & \phantom{$-$}0.054 {\scriptsize \phantom{$-$}[0.021, 0.089]} & 0.008 & 0.010 & 0.009 \\
\addlinespace[2pt]
Qwen-3-14B (max) & own answer & 0.068 {\scriptsize \phantom{$-$}[0.032, 0.104]} & \phantom{$-$}0.036 {\scriptsize \phantom{$-$}[0.005, 0.067]} & 0.001 & 0.071 & 0.003 \\
 & RFM first & 0.022 {\scriptsize [$-$0.008, 0.052]} & \phantom{$-$}0.024 {\scriptsize [$-$0.003, 0.053]} & 0.522 & 0.251 & 0.610 \\
 & RFM mean of 5 & 0.058 {\scriptsize \phantom{$-$}[0.020, 0.096]} & \phantom{$-$}0.026 {\scriptsize [$-$0.007, 0.058]} & 0.016 & 0.349 & 0.017 \\
\bottomrule
\end{tabular}

  \caption{Paired comparisons on the 500-sentence test split: the split probe (trained on the 1,000 RFM training sentences, pooling in brackets) minus the model's own answer, the first RFM concept vector and the mean of 5 (both with the frozen threshold). Differences and 95\% CIs from a paired bootstrap, and $p$-values from the paired bootstrap (Acc, AUC) and McNemar's test, each Holm-adjusted over the 12 model-pillar cells of its comparison.}
  \label{tab:paired}
\end{table*}

\section{Label efficiency}
\label{app:labeleff}
\begin{figure*}[t]
  \centering
  \includegraphics[width=\textwidth]{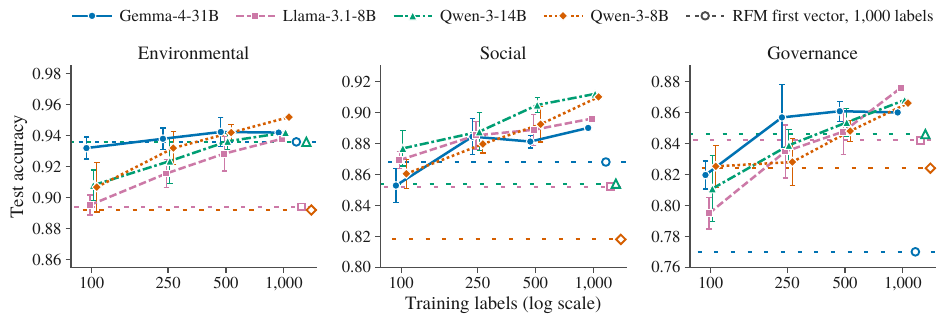}
  \caption{Three line charts, one per ESG pillar, of the split probe's test accuracy (500 test sentences) against the number of training labels (100, 250, 500 and 1,000, log scale), one line per model. Points below 1,000 labels are means over 5 random subsamples with error bars of one standard deviation. The 1,000-label point is the single fit on the full training split. A dashed horizontal line ending in a hollow marker at the right edge of each panel gives the same model's RFM first-vector accuracy with 1,000 training labels and the threshold frozen on validation. Each panel has its own y range, all of the same height (0.13 accuracy).}
  \label{fig:label-efficiency}
\end{figure*}

Table \ref{tab:label_eff} gives the values plotted in Figure \ref{fig:label-efficiency}.

\begin{table*}[tp]
  \centering
  \footnotesize
  \begin{tabular}{@{}lcccccc@{}}
\toprule
 & \multicolumn{4}{c}{Linear probe, training labels} & RFM first & Own \\
\cmidrule(lr){2-5}
Probe & 100 & 250 & 500 & 1,000 & 1,000 labels & answer \\
\midrule
\multicolumn{7}{l}{\textit{Environmental}} \\
Gemma-4-31B-it (last) & 0.932 {\scriptsize $\pm$0.007} & 0.938 {\scriptsize $\pm$0.007} & 0.942 {\scriptsize $\pm$0.009} & 0.942 & 0.936 & 0.930 \\
Llama-3.1-8B (last) & 0.895 {\scriptsize $\pm$0.006} & 0.916 {\scriptsize $\pm$0.009} & 0.928 {\scriptsize $\pm$0.011} & 0.938 & 0.894 & 0.864 \\
Qwen-3-8B (last) & 0.907 {\scriptsize $\pm$0.016} & 0.932 {\scriptsize $\pm$0.011} & 0.942 {\scriptsize $\pm$0.005} & 0.952 & 0.892 & 0.920 \\
Qwen-3-14B (last) & 0.908 {\scriptsize $\pm$0.010} & 0.924 {\scriptsize $\pm$0.014} & 0.936 {\scriptsize $\pm$0.005} & 0.942 & 0.936 & 0.916 \\
\midrule
\multicolumn{7}{l}{\textit{Social}} \\
Gemma-4-31B-it (mean) & 0.853 {\scriptsize $\pm$0.011} & 0.884 {\scriptsize $\pm$0.012} & 0.881 {\scriptsize $\pm$0.004} & 0.890 & 0.868 & 0.804 \\
Llama-3.1-8B (max) & 0.869 {\scriptsize $\pm$0.005} & 0.885 {\scriptsize $\pm$0.009} & 0.889 {\scriptsize $\pm$0.010} & 0.896 & 0.852 & 0.822 \\
Qwen-3-8B (max) & 0.860 {\scriptsize $\pm$0.009} & 0.880 {\scriptsize $\pm$0.006} & 0.892 {\scriptsize $\pm$0.011} & 0.910 & 0.818 & 0.836 \\
Qwen-3-14B (max) & 0.877 {\scriptsize $\pm$0.011} & 0.888 {\scriptsize $\pm$0.012} & 0.905 {\scriptsize $\pm$0.005} & 0.912 & 0.854 & 0.810 \\
\midrule
\multicolumn{7}{l}{\textit{Governance}} \\
Gemma-4-31B-it (max) & 0.820 {\scriptsize $\pm$0.009} & 0.857 {\scriptsize $\pm$0.021} & 0.861 {\scriptsize $\pm$0.006} & 0.860 & 0.770 & 0.756 \\
Llama-3.1-8B (last) & 0.795 {\scriptsize $\pm$0.010} & 0.835 {\scriptsize $\pm$0.017} & 0.848 {\scriptsize $\pm$0.015} & 0.876 & 0.842 & 0.820 \\
Qwen-3-8B (max) & 0.825 {\scriptsize $\pm$0.014} & 0.828 {\scriptsize $\pm$0.015} & 0.848 {\scriptsize $\pm$0.007} & 0.866 & 0.824 & 0.842 \\
Qwen-3-14B (max) & 0.811 {\scriptsize $\pm$0.021} & 0.839 {\scriptsize $\pm$0.010} & 0.854 {\scriptsize $\pm$0.009} & 0.868 & 0.846 & 0.800 \\
\bottomrule
\end{tabular}

  \caption{Test accuracy of the split probe (pooling in brackets) against the number of training labels, mean $\pm$ sd over 5 random subsets of the 1,000 RFM training sentences (one fit at 1,000). Reference columns: the first RFM concept vector trained on all 1,000 labels (threshold frozen on validation) and the model's own answer, on the same 500 test sentences.}
  \label{tab:label_eff}
\end{table*}

\section{Mixture test}
\label{app:mixture}
Table \ref{tab:mixture_rho} gives the per-cell correlations of the mixture test at $n = 10$.
On Social, Gemma-4-31B-it's primary probe is its mean-pooled probe, so the first two methods share that model's curve in Figure \ref{fig:mixture}.
On mixed paragraphs only ($0 < k < n$), the ranges are 0.76 to 0.91, 0.50 to 0.83 and 0.15 to 0.79 for the mean-pooled probe, the primary probe and the RFM mean of five.
With $n = 5$ the picture is the same, with ranges of 0.77 to 0.90, 0.72 to 0.86 and 0.64 to 0.83, and the RFM mean of five falls below the mean-pooled probe in eleven of twelve cells and below the primary probe in nine.

\begin{table*}[tp]
  \centering
  \footnotesize
  \begin{tabular}{@{}lcccccc@{}}
\toprule
 & \multicolumn{2}{c}{Probe, mean pooling} & \multicolumn{2}{c}{Primary probe} & \multicolumn{2}{c}{RFM, mean of 5} \\
\cmidrule(lr){2-3} \cmidrule(lr){4-5} \cmidrule(lr){6-7}
Model (pooling) & $\rho$ [95\% CI] & $\rho_{0<k<n}$ & $\rho$ [95\% CI] & $\rho_{0<k<n}$ & $\rho$ [95\% CI] & $\rho_{0<k<n}$ \\
\midrule
\multicolumn{7}{l}{\textit{Environmental}} \\
Gemma-4-31B-it (last) & 0.934 {\scriptsize [0.916, 0.946]} & 0.908 & 0.803 {\scriptsize [0.755, 0.841]} & 0.710 & 0.408 {\scriptsize [0.307, 0.503]} & 0.152 \\
Llama-3.1-8B (last) & 0.896 {\scriptsize [0.871, 0.915]} & 0.859 & 0.701 {\scriptsize [0.630, 0.760]} & 0.623 & 0.631 {\scriptsize [0.553, 0.699]} & 0.521 \\
Qwen-3-8B (last) & 0.909 {\scriptsize [0.884, 0.926]} & 0.876 & 0.794 {\scriptsize [0.738, 0.838]} & 0.687 & 0.696 {\scriptsize [0.635, 0.747]} & 0.611 \\
Qwen-3-14B (last) & 0.904 {\scriptsize [0.880, 0.921]} & 0.868 & 0.736 {\scriptsize [0.675, 0.788]} & 0.650 & 0.833 {\scriptsize [0.792, 0.865]} & 0.786 \\
\midrule
\multicolumn{7}{l}{\textit{Social}} \\
Gemma-4-31B-it (mean) & 0.890 {\scriptsize [0.863, 0.911]} & 0.831 & 0.890 {\scriptsize [0.863, 0.911]} & 0.831 & 0.692 {\scriptsize [0.625, 0.749]} & 0.614 \\
Llama-3.1-8B (max) & 0.868 {\scriptsize [0.837, 0.893]} & 0.813 & 0.797 {\scriptsize [0.741, 0.843]} & 0.729 & 0.814 {\scriptsize [0.774, 0.847]} & 0.749 \\
Qwen-3-8B (max) & 0.888 {\scriptsize [0.862, 0.907]} & 0.837 & 0.763 {\scriptsize [0.700, 0.816]} & 0.696 & 0.697 {\scriptsize [0.634, 0.750]} & 0.630 \\
Qwen-3-14B (max) & 0.852 {\scriptsize [0.818, 0.881]} & 0.802 & 0.863 {\scriptsize [0.828, 0.891]} & 0.807 & 0.749 {\scriptsize [0.696, 0.792]} & 0.667 \\
\midrule
\multicolumn{7}{l}{\textit{Governance}} \\
Gemma-4-31B-it (max) & 0.849 {\scriptsize [0.813, 0.877]} & 0.774 & 0.753 {\scriptsize [0.705, 0.793]} & 0.667 & 0.708 {\scriptsize [0.644, 0.763]} & 0.604 \\
Llama-3.1-8B (last) & 0.832 {\scriptsize [0.792, 0.864]} & 0.762 & 0.713 {\scriptsize [0.653, 0.762]} & 0.619 & 0.768 {\scriptsize [0.720, 0.808]} & 0.678 \\
Qwen-3-8B (max) & 0.823 {\scriptsize [0.784, 0.854]} & 0.758 & 0.608 {\scriptsize [0.526, 0.682]} & 0.501 & 0.695 {\scriptsize [0.634, 0.746]} & 0.604 \\
Qwen-3-14B (max) & 0.839 {\scriptsize [0.801, 0.869]} & 0.766 & 0.696 {\scriptsize [0.636, 0.748]} & 0.623 & 0.774 {\scriptsize [0.728, 0.813]} & 0.696 \\
\bottomrule
\end{tabular}

  \caption{Synthetic mixture test. Paragraphs of $n = 10$ sentences contain $k$ concept sentences ($0 \le k \le 10$, 330 paragraphs per pillar, model and method). $\rho$: Spearman correlation of the paragraph score with $k$, 95\% bootstrap CI. $\rho_{0<k<n}$: the same on mixed paragraphs only, excluding $k = 0$ and $k = n$. Primary probe: the model's pooling with the highest validation accuracy, named in brackets (identical to mean pooling for Gemma-4-31B-it on Social).}
  \label{tab:mixture_rho}
\end{table*}

\section{Score distributions}
\label{app:distributions}
\begin{figure*}[t]
  \centering
  \includegraphics[width=\textwidth]{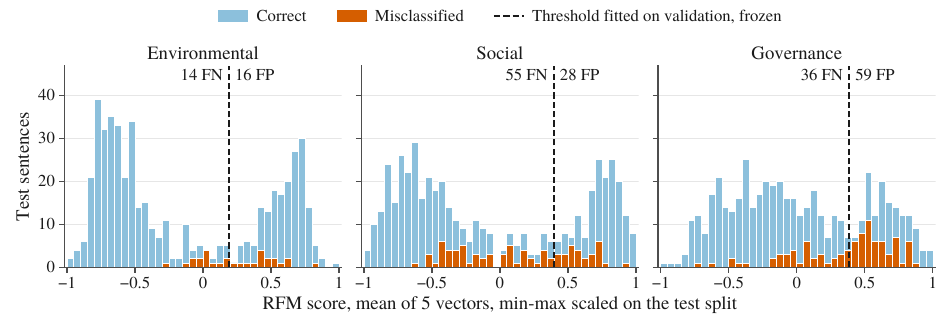}
  \caption{Distributions of the continuous RFM scores (Gemma-4-31B-it, mean of the top 5 concept vectors) on the 500 test sentences, min-max scaled to [-1, 1] within each ESG pillar. Left to right: Environmental, Social, Governance. Light blue denotes correctly classified sentences and the orange-red bars on the baseline misclassified ones. The dashed line marks the decision threshold, fitted on the validation split and applied frozen, with the numbers of false negatives and false positives on either side.}
  \label{fig:score-distributions}
\end{figure*}

In Figure \ref{fig:score-distributions} we can see the distribution of the scaled continuous scores from the RFM mean of top 5 concept vectors method scaled between -1 and 1 on all the ESG pillars.
The RFM mean-of-five score of Gemma-4-31B-it has a threshold fitted on the validation split and applied frozen, everything left of the threshold is classified as not containing the concept and right of the threshold it is.

Figure \ref{fig:score-distributions} (left) shows the distribution of Environmental scores.
The correctly classified sentences, in light blue, fall into two separate groups: the label 0 scores cluster around -0.63 and the label 1 scores cluster around 0.63 (medians).
The misclassified sentences, in orange-red, have a median of 0.26 and an interquartile range of 0.02 to 0.46, with 16 false positives and 14 false negatives against a threshold of 0.191.

Figure \ref{fig:score-distributions} (centre) shows a clear separation in the correctly classified sentences, in light blue, with label 0 scores clustering around -0.61 and label 1 scores clustering around 0.80 (medians).
The misclassified sentences, in orange-red, are spread between the two groups (median 0.11, interquartile range -0.28 to 0.52) and are mostly false negatives: 55, against 28 false positives, with a threshold of 0.397.

Figure \ref{fig:score-distributions} (right) shows the distribution of Governance scores.
Here there is no clear separation in the correctly classified sentences, in light blue (medians -0.28 and 0.68).
The misclassified sentences, in orange-red, lie predominantly above the threshold of 0.385, with 59 false positives and 36 false negatives (median 0.46, interquartile range 0.09 to 0.60).

Beyond classification, the RFM method provides a continuous score that may indicate how strongly a concept is present in a new sentence, though in the mixture test the mean-pooled probe's score tracks the number of concept sentences in a paragraph more closely (Section \ref{sec:mixture}).
The distributions of the RFM scores (Figure \ref{fig:score-distributions}) separate the correctly classified sentences most clearly on the Environmental pillar and least clearly on Governance, which matches the accuracy achieved on each pillar.
The optimal threshold is never exactly 0, but this says nothing about how well the concept vectors align with the concept.
Because the scores are min-max scaled within each pillar, 0 is the midpoint of the observed range rather than the point of zero similarity, so its position carries no interpretation of its own.
Reading the threshold as a measure of alignment would require the unscaled cosine similarities, which Appendix \ref{app:raw} reports.

The misclassified scores behave differently across pillars.
On Governance, the misclassified sentences have a median of 0.46 (interquartile range 0.09 to 0.60) and are mainly false positives.
On Social, the misclassified scores spread between the correctly labelled sentences (median 0.11, interquartile range -0.28 to 0.52) and are mostly false negatives, which suggests the similarity is uncertain on these sentences rather than confidently wrong.

\section{Raw cosine similarities}
\label{app:raw}
Table \ref{tab:raw_cosine} reports the raw cosine similarities behind the scaled RFM scores.

\begin{table*}[tp]
  \centering
  \footnotesize
  \begin{tabular}{@{}lrrcrrc@{}}
\toprule
 & \multicolumn{3}{c}{First concept vector} & \multicolumn{3}{c}{Mean of 5} \\
\cmidrule(lr){2-4} \cmidrule(lr){5-7}
Model & Mean & SD & AUC & Mean & SD & AUC \\
\midrule
\multicolumn{7}{l}{\textit{Environmental}} \\
Gemma-4-31B-it & $-$0.0089 & 0.0081 & 0.975 & 0.0244 & 0.0081 & 0.983 \\
Llama-3.1-8B & $-$0.0352 & 0.0189 & 0.962 & 0.0180 & 0.0106 & 0.943 \\
Qwen-3-8B & $-$0.0037 & 0.0083 & 0.965 & $-$0.0064 & 0.0050 & 0.912 \\
Qwen-3-14B & $-$0.0101 & 0.0052 & 0.984 & 0.0223 & 0.0033 & 0.980 \\
\midrule
\multicolumn{7}{l}{\textit{Social}} \\
Gemma-4-31B-it & $-$0.0078 & 0.0051 & 0.943 & $-$0.0006 & 0.0066 & 0.922 \\
Llama-3.1-8B & $-$0.0070 & 0.0202 & 0.922 & 0.0099 & 0.0127 & 0.902 \\
Qwen-3-8B & 0.0338 & 0.0073 & 0.907 & 0.0034 & 0.0042 & 0.841 \\
Qwen-3-14B & 0.0083 & 0.0057 & 0.935 & 0.0292 & 0.0047 & 0.910 \\
\midrule
\multicolumn{7}{l}{\textit{Governance}} \\
Gemma-4-31B-it & $-$0.0130 & 0.0053 & 0.811 & $-$0.0042 & 0.0077 & 0.855 \\
Llama-3.1-8B & $-$0.0522 & 0.0199 & 0.877 & $-$0.0030 & 0.0134 & 0.881 \\
Qwen-3-8B & $-$0.0457 & 0.0067 & 0.863 & 0.0098 & 0.0043 & 0.847 \\
Qwen-3-14B & $-$0.0408 & 0.0057 & 0.872 & $-$0.0234 & 0.0033 & 0.871 \\
\bottomrule
\end{tabular}

  \caption{Raw cosine similarity between a sentence's activations and the concept vector, before min-max scaling, on the 500 test sentences: mean, standard deviation and the AUC of the raw score (scaling is monotone, so the AUC equals the RFM AUC). Means and SDs to four decimals because raw cosines are of order 0.01.}
  \label{tab:raw_cosine}
\end{table*}

\section{Background details}
\label{app:background}

\subsection{Activations and concept vectors}
Modern decoder-only LLMs utilise the transformer architecture introduced by \citet{NIPS2017_3f5ee243}: a stack of identical computational blocks, also referred to as the hidden layers, each consisting of a self-attention layer followed by a feed-forward layer.
Attention in these models is causal: when a token's representation is updated, it can see itself and the tokens that come before it, never the ones that have yet to follow.
A tokenised prompt passes through the stack once, and every block leaves one vector per token, the hidden state or activation, in that block's activation space, a high-dimensional space typically of several thousand dimensions.
Depth is model-specific, with some models containing 30 blocks or fewer and others 80 or more.
These activations are the model's internal representation of what it has read up until that point \citep{10.1162/coli_a_00422}.
Each block has its own activation space and encodes different kinds of information.
Early blocks tend to capture low-level features, while middle to late layers capture more abstract ones \citep{tenney-etal-2019-bert}.

The linear representation hypothesis holds that concepts are accessible in this space as directions \citep{pmlr-v235-park24c,marks2024geometry}.
A concept vector is such a direction in the activation space of a single layer.
When correctly found, it should point towards the features that matter most for predicting the concept.
A concept vector can be used in two ways: to steer a model toward the concept by adding the vector to new activations, or to monitor how active the concept is within new activations. This paper only monitors.

Monitoring is worth the trouble because a model knows more than it expresses in its response \citep{doi:10.1126/science.aea6792,burns2023discovering,azaria-mitchell-2023-internal}, so reading the activations recovers what the output leaves out.
\citet{gurnee2026workspace} sharpen this picture: only a small subspace of the activations, concentrated in the intermediate layers, carries the content the model can report verbally, while the rest drives processing the model never puts into words.
Monitoring is also cheap.
Verifying a model's truthfulness previously relied on separate models to check for hallucination and accuracy, whereas monitoring can run on a single GPU within a minute \citep{doi:10.1126/science.aea6792,zou2023transparency}.

\subsection{Linear probing and RFM}
Two supervised methods turn labelled activations into such a direction while leaving the model frozen, and they are the pair this paper compares.

Probes of this kind read graded quantities such as place and time directly out of the activations of a frozen model \citep{gurnee2024language}.

The Recursive Feature Machine (RFM) algorithm is a supervised model developed by \citet{radhakrishnan2024mechanism}.
It seeks to find a separation between classes by using non-linearity on the labelled activations of LLMs as input.
It does so by computing the Average Gradient Outer Product (AGOP) matrix, which weighs which features are most important for telling the classes apart.
This matrix is recalculated at the end of each iteration, and as the algorithm repeats, it keeps updating the matrix to capture which features matter most.
\citet{doi:10.1126/science.aea6792} found that this same AGOP matrix can also be used to extract concept vectors.
By applying eigendecomposition, the top $n$ eigenvectors are extracted from the matrix.
These vectors best distinguish between the classes, and they become the concept vectors.
Finding the concept vectors this way was not the original purpose of the RFM algorithm. It is a by-product of how the AGOP matrix captures the most important directions for separating the classes.

Both directions have already been put to work as measures: \citet{doi:10.1126/science.aea6792} monitor RFM concept vectors across hundreds of concepts, and \citet{yang2024llmmeasure} project the hidden states of a frozen LLM onto a learned concept direction to obtain continuous text-based measures for social science research.
Neither compares the two extractors against each other.

\subsection{Token pooling}
Both extractors usually require a single fixed-size vector per layer, but after a prompt of N tokens is processed by a block, the block outputs a sequence of N activation vectors, one per token.
Token pooling refers to the method used to transform that sequence into one vector, and three strategies are common: last-token activation, mean pooling, and max pooling \citep{reimers-gurevych-2019-sentence}.

Last-token activation uses the activation vector of the final token in the sequence as the prompt's representation.
Because tokens in a decoder-only LLM are causal, the last token has seen and been updated using the entire sequence of tokens that have preceded it, and can thus be interpreted as containing a summary of it.

Mean pooling takes the mean of the entire sequence of N activation vectors, producing a single vector.
It treats every token equally, which allows signals to remain visible but dilutes signals that are only active in a few tokens.

Max pooling takes the maximum value per signal out of the sequence of N token activation vectors, producing a single vector.
This emphasises the strongest activation along each direction, which strengthens signals that are only active in a few tokens, but also makes max pooling more sensitive to outliers than mean pooling.

On the Environmental task, last-token pooling performed best, while on the Social and Governance tasks max pooling was the strongest.
One possible explanation is that Environmental is the easiest task, as all activation-based methods achieve high results on it, so the last token alone may already carry a rich enough summary of the sentence.
The other two tasks appear more difficult, and max pooling may help by capturing the few strong signals that are present more effectively than the other strategies.

\section{Statistical procedures}
\label{app:stats}
In Tables \ref{tab:main}, \ref{tab:full_environmental}, \ref{tab:full_social} and \ref{tab:full_governance}, the activation-based and own-answer cells carry a 95\% bootstrap confidence interval over sentences, and the embedding, TF-IDF and fine-tuned rows the sd over folds.
Each interval runs from the 2.5th to the 97.5th percentile of 10,000 resamples with seed 0.
For the CV probe, the point estimate is the mean over the five folds, while the interval resamples the pooled out-of-fold predictions of all 2,000 sentences.
On the 500-sentence test split, we compare the split probe against the model's own answer and against RFM (first concept vector and mean of five, threshold frozen).
The comparisons use a paired bootstrap on the differences in accuracy and AUC and McNemar tests, each Holm-corrected over the twelve model-pillar cells of its comparison.
The paired bootstrap draws the same resamples for both methods and reports a two-sided p-value, and McNemar's test uses the exact binomial form.
Label-efficiency curves refit the split probe on 100, 250 and 500 training sentences drawn from the 1,000 (five stratified draws each) and on all 1,000, and test every fit on the same 500 test sentences.
The synthetic mixture test (Figure \ref{fig:mixture}) composes paragraphs of $n = 5$ and $n = 10$ sentences containing $k$ concept sentences, for $k$ from 0 to $n$, with 30 paragraphs per $k$ (180 and 330 paragraphs per pillar, model and method).
The sentences are drawn from the validation and test splits (sentences 1,000 to 1,999 after shuffling) with seed 0.
A sentence appears at most once within a paragraph, but can appear in several paragraphs.
The sentences of a paragraph are joined with a space, and each paragraph is then inserted whole into the wrapper and scored exactly like a single sentence, by the split probe and the RFM concept vectors trained on the 1,000 training sentences.
The probe scores a paragraph by the log-odds of its predicted probability and RFM by its raw cosine score.
Each method scores every paragraph, and the scores are standardised within method and pillar.
We report the Spearman correlation $\rho$ of the score with $k$, over all paragraphs and restricted to mixed paragraphs ($0 < k < n$).
The confidence intervals on $\rho$ come from 10,000 bootstrap resamples of the paragraphs. Because paragraphs share sentences, these intervals are somewhat narrower than they would be for independent paragraphs.

\section{Practical implications}
\label{app:practical}
Because our activation-based methods come close to the fully fine-tuned models of \citet{schimanski2024bridging} while being based on frozen, out-of-the-box LLMs, they offer a valid option for anyone who does not have enough data or computing power to fine-tune a model themselves.
The main saving will be in the gathering of data and training time, as with our method there is no domain-specific fine-tuning required.
The cost is small in absolute terms: on a single H100, extracting activations and fitting concept vectors for one 8B model and one pillar takes under four minutes, and the full linear probe with nested cross-validation takes about five.
Extracting the activations of an LLM still has a computing cost, and a small amount of data is still needed to extract the desired concept vector.

These considerations gain weight as corporate communication itself becomes machine-generated.
Generative models can already produce corporate text that readers struggle to tell apart from a CEO's own \citep{ChoudhuryEtAl2026}.
In such a landscape, what a text appears to say and what concepts it actually carries may drift apart, and a measure that reads concept content from a model's internal representations rather than from surface cues offers a complementary check.

For practitioners, the lightweight probe is the default choice: it is the strongest activation-based method we tested, matches the embedding baseline and is among the cheapest to run.
Whichever method is used, the wrapper deserves care: small changes to it move accuracy by two to three points on the 8B models and up to 5.6 points for the concept vectors on Gemma-4-31B-it, mostly inside the paired intervals, while leaving the concept undefined, as the bare sentence and the short question do, costs the concept vectors most of their accuracy range (Appendix A).

\end{document}